\documentclass[letterpaper]{article} 
\usepackage{aaai2026}  
\usepackage{times}  
\usepackage{helvet}  
\usepackage{courier}  
\usepackage[hyphens]{url}  
\usepackage{graphicx} 
\usepackage{natbib}  
\usepackage{caption} 
\usepackage{algorithm}
\usepackage{algorithmic}

\usepackage{newfloat}
\usepackage{listings}
\DeclareCaptionStyle{ruled}{labelfont=normalfont,labelsep=colon,strut=off} 
\floatstyle{ruled}
\newfloat{listing}{tb}{lst}{}
\floatname{listing}{Listing}
\usepackage{amsmath}
\usepackage{amssymb} 
\usepackage{tikz}    
\tikzstyle{block} = [rectangle, draw, fill=cyan,
    text width=7em, text centered, rounded corners, minimum height=2em]
\tikzstyle{arrow} = [thick,->,>=stealth]
\usetikzlibrary{automata, arrows.meta, positioning, shapes.geometric}
\usetikzlibrary{calc} 

\usepackage{subcaption}
\usepackage{array} 
\usepackage[thinlines]{easytable}  
\usepackage{stmaryrd} 
\usepackage{amsthm} 
\usepackage{microtype}

\usepackage{tabularx} 

\usepackage{todonotes}[]{}
\presetkeys{todonotes}{inline}{}

\usepackage{marvosym}
\usepackage[marvosym]{tikzsymbols}

\newtheorem{theorem}{Theorem}

\newtheorem{proposition}[theorem]{Proposition}

\newtheorem{definition}{Definition}

\newtheorem{assumption}{Assumption}

\iftrue
    \newcommand{\aneta}[1]{\textcolor{green}{Aneta: #1}}
    \newcommand{\jendrik}[1]{\textcolor{orange}{Jendrik: #1}}
    \newcommand{\kristina}[1]{\textcolor{cyan}{Kristina: #1}}
    \newcommand{\nikos}[1]{\textcolor{red}{Nikos: #1}}
    \newcommand{\thanos}[1]{\textcolor{yellow}{Thanos: #1}}
\else
    \newcommand{\aneta}[1]{}
    \newcommand{\jendrik}[1]{}
    \newcommand{\kristina}[1]{}
    \newcommand{\nikos}[1]{}
    \newcommand{\thanos}[1]{}
\fi

\newcommand{\props}{\ensuremath{{\mathcal{P}}}}
\newcommand{\psa}{\ensuremath{{\props\mathit{SA}}}}

\newcommand{\dec}{\hspace{-.05em}\raisebox{.15ex}{\footnotesize$\downarrow$}}

\newcommand{\interp}[1]{\llbracket #1 \rrbracket}

\newcommand{\set}[1]{\{#1\}}

\newcommand{\delete}[1]{}

\newcommand{\agenda}{\ensuremath{\mathcal{T}}}
\newcommand{\tasksubset}{\ensuremath{T}}
\newcommand{\task}{\tau}

\newcommand{\decbar}{$\dec$\hspace{-0.48em}\raisebox{-0.8ex}{--}\hspace{0em}}

\newcommand{\MP}{\ensuremath{\mathcal{P}}}
\newcommand{\MR}{\ensuremath{\mathcal{R}}}
\newcommand{\MD}{\ensuremath{\mathcal{D}}}

\newcommand{\EpiL}{\ensuremath{K}}
\newcommand{\EpiLExp}{\ensuremath{\bar{K}}}

\newcommand{\EpiLBest}{\ensuremath{\bar{K}_{\text{op}}}} 
\newcommand{\EpiLMin}{\ensuremath{\bar{K}_{\text{mi}}}} 

\title{Reinforcement Learning for Long-Horizon Unordered Tasks:\\ From Boolean to Coupled Reward Machines}
\author{
   Kristina Levina\textsuperscript{\rm 1,\rm 2},
   Nikolaos Pappas\textsuperscript{\rm 1},
   Athanasios Karapantelakis\textsuperscript{\rm 2},\\
   Aneta Vulgarakis Feljan\textsuperscript{\rm 2},
   Jendrik Seipp\textsuperscript{\rm 1}
}
\affiliations{
    \textsuperscript{\rm 1}Linköping University, Linköping, Sweden\\
    \textsuperscript{\rm 2}Ericsson Research, Stockholm, Sweden\\
   \{kristina.levina,
   nikolaos.pappas,
   jendrik.seipp\}@liu.se \\
   \{aneta.vulgarakis,
   athanasios.karapantelakis\}@ericsson.com
}

\nocopyright

\begin{document}

\maketitle

\begin{abstract}
Reward machines (RMs) inform reinforcement learning agents about the reward structure of the environment, enabling support for non-Markovian tasks and improving sample efficiency.
However, learning with RMs is ill-suited for long-horizon problems where subtasks can be completed in any order.
In such cases, the amount of information to learn increases exponentially with the number of unordered subtasks.
We address this issue by introducing three generalisations of RMs:
    (1) \emph{Numeric} RMs allow users to express complex tasks in a compact form.
    (2) In \emph{agenda} RMs, states are associated with an agenda that tracks the remaining subtasks to complete.
    (3) \emph{Coupled} RMs have coupled states associated with each subtask in the agenda.
In addition, we introduce QCoRM, a new task-decomposition $Q$-learning-based algorithm that leverages coupled RMs and preserves global optimality guarantees in tabular settings.
Our experiments across four domains---featuring both discrete and continuous action and state spaces---demonstrate that QCoRM scales better than baseline algorithms for long-horizon problems with unordered subtasks.
\end{abstract}


\section{Introduction}

Reinforcement learning (RL) agents learn policies---mappings from states to actions---by interacting with an environment and receiving reward-based feedback~\citep{sutton2018reinforcement}.
Designing effective reward functions to guide learning is often difficult and time-consuming~\citep{eschmann2021reward}.
Moreover, learning complex, long-horizon tasks is challenging because rewards are often sparse and delayed~\citep{dulac2021challenges}.
Task-decomposition methods mitigate these challenges by breaking down complex tasks into simpler subtasks~\citep{dietterich2000hierarchical}.

One class of task-decomposition methods leverages logical formulae, referred to as \emph{specifications}, to express the desired behaviour of the agent~\citep{krasowski2023provably}.
There are various specification languages and methods for compiling specifications into rewards~\citep{
camacho2019ltl}.

\begin{figure}[tpb]
\centering
\begin{subfigure}[]{0.09\textwidth}

    \begin{tikzpicture}
    \draw[step=0.5,gray] (0,0) grid (1.5,2);

    \draw[line width = 0.9 mm] (0,0) rectangle (1.5,2);
    \draw (0.6,1.71) node[anchor=east] {\texttt{b}$_1$};
    \draw (0.6,0.24) node[anchor=east] {\texttt{b}$_2$};
    \draw (1,1.74) node[anchor=east] {\texttt{s}};
    \draw (1.5,1.24) node[anchor=east] {\texttt{A}};
    \end{tikzpicture}
    \caption{}
    \label{fig:delivery}
\end{subfigure}
\hfill
\begin{subfigure}[]{0.36\textwidth}
    \begin{tikzpicture}[node distance = 2.5cm, on grid, auto]
        {\small
        \node (u0) [state, initial, initial text = {}, minimum size=16pt] {$u_0$};
        \node (u1) [state, right = of u0, minimum size=16pt] {$u_1$};
        \node (u2) [state, accepting, right = of u1, minimum size=16pt] {$u_2$};

        \path [-stealth, thick]
            (u0) edge   [loop above]    node {$\langle$$b$!; $0$$\rangle$}   (u0)
            (u0) edge   [bend right]    node[below] {$\langle$$b$$\dec$$\lor$$b$\decbar; $0$$\rangle$}   (u1)
            (u1) edge   [loop above]    node {$\langle$$\lnot \texttt{s}$; $0$$\rangle$}   (u1)
            (u1) edge                   node[above right, pos = 0.05]  {$\langle$$\texttt{s}$$\land$$b$\decbar; $1$$\rangle$}   (u2)
            (u1) edge   [bend right]     node[above]  {$\langle$$\texttt{s}$$\land$$b$!; $0$$\rangle$}   (u0)
            ;
    }
    \end{tikzpicture}
    \caption{}
    \label{fig:rm-num}
\end{subfigure}
\hfill
 \caption{(a) Example Delivery instance with agent \texttt{A}, station \texttt{s}, and two boxes
 \label{fig:delivery-task-and-numeric-rm}
 \texttt{b}$_1$ and \texttt{b}$_2$.
 (b) Numeric reward machine (RM) for the Delivery domain.
 A discrete numeric variable counts the number of uncollected boxes and is mapped to numeric feature $b$.
 Here, $b\dec$ is true when a box is collected,
 $b\decbar$ is true when all boxes are collected,
 and $b$! is true when not all boxes are collected.
 Boolean feature $\texttt{s}$ is true when the agent arrives at the station.
 The agent receives a reward of $1$ upon delivering all boxes to the station and $0$ otherwise.
 }

\end{figure}

Another approach is to specify tasks in abstract graphs.
\citet{icarte2022reward} introduced Boolean reward machines (RMs)---automata in which states encode task progression.
Augmenting environment states with RM states enables support for non-Markovian tasks.
RMs have been shown to outperform plain reward functions and specifications in many settings~\citep{neary2020reward,unniyankal2023rmlgym}.

Although existing RM-aware algorithms improve sample efficiency, complex, long-horizon tasks remain challenging---especially when subtasks can be completed in any order.
In such cases, the number of RM states grows exponentially with the number of subtasks: one state per subset of completed items, yielding $2^N$ states for $N$ subtasks.
Consequently, the agent must explore up to $N!$ distinct completion orderings to discover an optimal one.
The resulting exponential state space makes learning difficult, and modelling an RM with so many states becomes tedious and error-prone.

\begin{figure}[t!]
\centering
\begin{subfigure}[]{0.45\textwidth}
\begin{tikzpicture}[node distance=1.7cm, on grid, auto]
{\small
    \node[state, initial, initial text = {}, minimum size=16pt] (s0) {$u_0$};
    \node[state, right=of s0, yshift=0.6cm, minimum size=16pt] (s1) {$u_1$};
    \node[state, below=of s1, minimum size=16pt, yshift=0.4cm] (s2) {$u_2$};
    \node[state, right=of s1, minimum size=16pt] (s3) {$u_3$};
    \node[state, right=of s2, minimum size=16pt] (s4) {$u_4$};
    \node[state, right=of s3, minimum size=16pt, fill = yellow] (s5) {$u_5$};
    \node[state, right=of s4, minimum size=16pt, fill = yellow] (s6) {$u_6$};
    \node[state, accepting, right=of s5, minimum size=16pt, fill = cyan] (s7) {$u_7$};
    \node[state, accepting, right=of s6, minimum size=16pt, fill = cyan] (s8) {$u_8$};

    \path [-stealth, thick]
        (s0) edge node[sloped] {$\langle$\texttt{b}$_1$; $0$$\rangle$} (s1)
        (s0) edge node[sloped] {$\langle$\texttt{b}$_2$; $0$$\rangle$} (s2)
        (s1) edge node[sloped] {$\langle$\texttt{s}; $0$$\rangle$} (s3)
        (s2) edge node[sloped] {$\langle$\texttt{s}; $0$$\rangle$} (s4)
        (s3) edge node {$\langle$\texttt{b}$_2$; $0$$\rangle$} (s5)
        (s4) edge node {$\langle$\texttt{b}$_1$; $0$$\rangle$} (s6)
        (s5) edge node[sloped] {$\langle$\texttt{s}; $1$$\rangle$} (s7)
        (s6) edge node[sloped] {$\langle$\texttt{s}; $1$$\rangle$} (s8)

        (s0) edge   [loop below]    node {$\langle$$\lnot$\texttt{b}$_1$$\land$$\lnot$\texttt{b}$_2$; $0$$\rangle$}   (s0)
        (s1) edge   [loop above]    node {$\langle$$\lnot$\texttt{s}; $0$$\rangle$}   (s1)
        (s2) edge   [loop below]    node {$\langle$$\lnot$\texttt{s}; $0$$\rangle$}   (s2)
        (s5) edge   [loop above]    node {$\langle$$\lnot$\texttt{s}; $0$$\rangle$}   (s5)
        (s6) edge   [loop below]    node {$\langle$$\lnot$\texttt{s}; $0$$\rangle$}   (s6)
        (s3) edge   [loop above]    node {$\langle$$\lnot$\texttt{b}$_2$; $0$$\rangle$}   (s3)
        (s4) edge   [loop below]    node {$\langle$$\lnot$\texttt{b}$_1$; $0$$\rangle$}   (s4)

        ;
}
\end{tikzpicture}
\caption{Boolean RM.}
\label{fig:rm-bool}
\end{subfigure}
\begin{subfigure}[]{0.45\textwidth}
\centering
\begin{tikzpicture}[node distance=1.6cm, on grid, auto]
 \tikzset{
        myState/.style={
            state,
            minimum size=15pt,
            shape=rectangle,
            rounded corners=5pt
        }
    }
{\fontsize{8.5}{10}
    \node[myState, initial, initial text = {}] (s0) {$0$\set{$1$,$2$}\set{$1$,$2$}};
    \node[myState, right=of s0, yshift=0.7cm, xshift=0.3cm] (s1) {$1$\set{$2$}s};
    \node[myState, below=of s1, yshift = 0.25cm] (s2) {$1$\set{$1$}s};
    \node[myState, right=of s1] (s3) {$2$\set{$2$}$2$};
    \node[myState, right=of s2] (s4) {$2$\set{$1$}$1$};
    \node[myState, right=of s3, yshift=-0.7cm, fill = yellow] (s5) {$3$\set{}s};
    \node[myState, accepting, right=of s5, fill = cyan, xshift = -0.2cm] (s6) {$4$\set{}};

    \path [-stealth, thick]
        (s0) edge node[sloped] {$\langle$\texttt{b}$_1$; $0$$\rangle$} (s1)
        (s0) edge node[sloped,below] {$\langle$\texttt{b}$_2$; $0$$\rangle$} (s2)
        (s1) edge node[sloped] {$\langle$\texttt{s}; $0$$\rangle$} (s3)
        (s2) edge node[sloped] {$\langle$\texttt{s}; $0$$\rangle$} (s4)
        (s3) edge node[sloped] {$\langle$\texttt{b}$_2$; $0$$\rangle$} (s5)
        (s4) edge node[sloped] {$\langle$\texttt{b}$_1$; $0$$\rangle$} (s5)
        (s5) edge node[sloped] {$\langle$\texttt{s}; $1$$\rangle$} (s6)

        (s0) edge   [loop above]    node {$\langle$$\lnot$\texttt{b}$_1$$\land$$\lnot$\texttt{b}$_2$; $0$$\rangle$}   (s0)
        (s1) edge   [loop above]    node {$\langle$$\lnot$\texttt{s}; $0$$\rangle$}   (s1)
        (s2) edge   [loop below]    node {$\langle$$\lnot$\texttt{s}; $0$$\rangle$}   (s2)
        (s5) edge   [loop above]    node {$\langle$$\lnot$\texttt{s}; $0$$\rangle$}   (s5)
        (s3) edge   [loop above]    node {$\langle$$\lnot$\texttt{b}$_2$; $0$$\rangle$}   (s3)
        (s4) edge   [loop below]    node {$\langle$$\lnot$\texttt{b}$_1$; $0$$\rangle$}   (s4)

        ;
}
\end{tikzpicture}
\caption{Agenda RM.
}
\label{fig:rm-sem}
\end{subfigure}
\begin{subfigure}[]{0.45\textwidth}
\begin{tikzpicture}[node distance=1.6cm, on grid, auto]
{\fontsize{8.5}{10}
\tikzset{
        myState/.style={
            state,
            minimum size=16pt,
            shape=rectangle,
            rounded corners=5pt
        }
    }

    \node[myState, initial, initial text = {}, minimum size=16pt] (s01) {$0$\set{$1$,$2$}$1$};
    \node[myState, below=of s01, initial, initial text = {}, minimum size=16pt, yshift = 0.3cm] (s02) {$0$\set{$1$,$2$}$2$};
    \node[myState, right=1.9cm of s01, minimum size=16pt] (s1) {$1$\set{$2$}s};
    \node[myState, below=of s1, minimum size=16pt, yshift = 0.3cm] (s2) {$1$\set{$1$}s};
    \node[myState, right=of s1, minimum size=16pt] (s3) {$2$\set{$2$}$2$};
    \node[myState, right=of s2, minimum size=16pt] (s4) {$2$\set{$1$}$1$};
    \node[myState, right=of s3, minimum size=16pt, yshift=-0.7cm, fill = yellow] (s5) {$3$\set{}s};
    \node[myState, accepting, right=of s5, minimum size=16pt, fill = cyan] (s6) {$4$\set{}};

      \draw[red, thick, dashed] ($(s01)!0.5!(s02)$) ellipse[x radius=0.8cm, y radius=1.4cm];

    \path [-stealth, thick]
        (s01) edge node[sloped] {$\langle$\texttt{b}$_1$; $0$$\rangle$} (s1)
        (s02) edge node[sloped] {$\langle$\texttt{b}$_2$; $0$$\rangle$} (s2)
        (s1) edge node[sloped] {$\langle$\texttt{s}; $0$$\rangle$} (s3)
        (s2) edge node[sloped] {$\langle$\texttt{s}; $0$$\rangle$} (s4)
        (s3) edge node[sloped] {$\langle$\texttt{b}$_2$; $0$$\rangle$} (s5)
        (s4) edge node[sloped] {$\langle$\texttt{b}$_1$; $0$$\rangle$} (s5)
        (s5) edge node[sloped] {$\langle$\texttt{s}; $1$$\rangle$} (s6)

        (s01) edge   [loop above]    node {$\langle$$\lnot$\texttt{b}$_1$; $0$$\rangle$}   (s01)
        (s02) edge   [loop below]    node {$\langle$$\lnot$\texttt{b}$_2$; $0$$\rangle$}   (s02)
        (s1) edge   [loop above]    node {$\langle$$\lnot$\texttt{s}; $0$$\rangle$}   (s1)
        (s2) edge   [loop below]    node {$\langle$$\lnot$\texttt{s}; $0$$\rangle$}   (s2)
        (s5) edge   [loop above]    node {$\langle$$\lnot$\texttt{s}; $0$$\rangle$}   (s5)
        (s3) edge   [loop above]    node {$\langle$$\lnot$\texttt{b}$_2$; $0$$\rangle$}   (s3)
        (s4) edge   [loop below]    node {$\langle$$\lnot$\texttt{b}$_1$; $0$$\rangle$}   (s4)

        ;
}
\end{tikzpicture}
\begin{subtable}[b]{0.45\textwidth}
{\fontsize{8.5}{10}
    \centering
    \setlength{\tabcolsep}{2.35pt}
    \begin{tabular}{|c|c|c|c|c|c|c|c|c|}
        \hline
        \textbf{State} & $0$\set{$1$,$2$}$1$ & $0$\set{$1$,$2$}$2$ & $1$\set{$1$}s & $1$\set{$2$}s & $2$\set{$1$}$1$  & $2$\set{$2$}$2$ & $3$\set{}s & $4$\set{} \\ \hline
        $\eta$         & $12$   & $10$  & $6$   & $9$  & $2$  &  $8$ & $1$ & $0$ \\ \hline
    \end{tabular}
}
\end{subtable}
\caption{Coupled RM.
The red dashed line indicates that the agent is in states $0$\set{$1$,$2$}$1$ and $0$\set{$1$,$2$}$2$ concurrently.
The $\eta$ values were found using $Q$-learning with the coupled RM (QCoRM).
}\label{fig:rm-coupled}
\end{subfigure}
\caption{RMs for the Delivery task with two boxes shown in Figure~\ref{fig:delivery}.
Features \texttt{b}$_1$ and \texttt{b}$_2$ become true when the agent collects boxes $1$ and $2$, respectively.
Symmetric states are shown in yellow and blue.}
\end{figure}

We propose three generalisations of RMs that address the challenges mentioned above: \emph{numeric}, \emph{agenda}, and \emph{coupled} RMs.
To illustrate them, we introduce a running example based on a variant of the Delivery domain---a deterministic finite-grid world with four-connected cells, some containing boxes.
The agent must collect all boxes and deliver them to a designated station cell.
Entering a box cell automatically collects the box, and the agent cannot accidentally drop it, but only one box can be carried at a time.
Figure~\ref{fig:delivery} illustrates an example environment with two boxes.

First, we introduce numeric features to RMs to simplify RM design for the user.
For compact modelling, unordered subtasks are encapsulated into a discrete numeric variable, which is then mapped to a numeric feature.
For our running example, Figure~\ref{fig:rm-num} shows a numeric RM for two boxes.
Numeric variable $w$ with domain $\mathcal{D}_{w} = \set{0, 1, 2}$ counts the number of uncollected boxes.
However, the agent cannot directly use this RM for learning because RM states $u_0$ and $u_1$
are visited twice---once for each collected box.
As a solution, the given numeric RM can be translated to a Boolean RM (Figure~\ref{fig:rm-bool}).
However, Boolean RMs scale poorly: the number of states grows exponentially with the number of boxes, and the agent must learn exponential information.

To solve these problems, we label each RM state $u$ with $\langle d, \tasksubset, x \rangle$, where $d$ is the distance from the initial state in the RM to $u$; $\tasksubset$ is an \emph{agenda}, a set of remaining subtasks to be completed in any order; and $x$ is the current objective.
For conciseness, we abbreviate $\langle d, \{\task_1, \task_2, \dots\}, x \rangle$ with $d\{\task_1, \task_2, \dots\}x$.
States with identical labels are symmetric---they share a single $Q$-function---and can be merged.
For example, in the agenda RM in Figure~\ref{fig:rm-sem}, yellow state $3$\{\}s corresponds to states $u_5$ and $u_6$ in the Boolean RM.
State reduction accelerates learning by decreasing the number of distinct paths the agent must explore in the RM.
Moreover, associating states with an agenda $\tasksubset$ allows us to split states by the subtasks in $\tasksubset$, yielding a \emph{coupled} RM (Figure~\ref{fig:rm-coupled}).
For example, if $\tasksubset$ = \set{$\task_1$,$\task_2$}, the state with current objective $x = \tasksubset$ can be split into two coupled states: $d$\set{$\task_1$,$\task_2$}$\task_1$ and $d$\set{$\task_1$,$\task_2$}$\task_2$.
Conceptually, the agent occupies all coupled states concurrently and may transition from any of them.

We exploit the structure of coupled RMs in a new task-decomposition $Q$-learning-based algorithm: QCoRM.
In QCoRM, the agent learns low-level policies for completing individual subtasks along with a high-level policy over coupled RM states to decide completion order.
The algorithm leverages the fact that each state in a coupled RM always corresponds to a single subtask.
For the high-level policy, we treat coupled RM states as multi-armed bandits.
As the decision metric, we use the expected number of steps $\eta(u)$ from each RM state $u$ to a goal state.
The high-level policy coordinates low-level learning by shaping the reward signals using the episode length, preserving global optimality in tabular settings.
In our running example, the QCoRM agent learns low-level policies to locate box $1$, box $2$, and the station, as indicated by the right parts of the RM-state labels: $1$, $2$, and s, respectively.
In coupled states $0$\set{$1$,$2$}$1$ and $0$\set{$1$,$2$}$2$, the agent decides to transition from $0$\set{$1$,$2$}$2$ because its $\eta$ value is $10$, which is lower than $12$ for state $0$\set{$1$,$2$}$1$.

While we focus on unordered subtasks,
QCoRM can reduce the number of low-level policies to learn in sequential tasks.
For example, to reach \texttt{a}, \texttt{b}, and \texttt{a} in order (RM states $u_0$, $u_1$, and $u_2$), the QCoRM agent learns two reusable policies to reach \texttt{a} and \texttt{b}.
In contrast, the Boolean-RM agent learns three policies, one per RM state $u_0$, $u_1$, and $u_2$.

\section{Background}

We begin by providing background information on RL and RMs. For more details on these topics, we refer to~\citet{sutton2018reinforcement} and~\citet{icarte2022reward}, respectively.

\subsection{Reinforcement Learning}

Single-agent RL tasks are formalised via Markov decision processes (MDPs), defined by the tuple $M = \langle S, s_0, A, p, r, \gamma \rangle$, where $S$ is a finite set of environment states, $s_0 \in S$ is an initial state, $A$ is a finite set of actions, $p : S \times A \to \Delta(S)$ is a transition probability function, $\Delta(S)$ is the probability simplex over $S$, $r : S \times A \times S \to \mathbb{R}$ is a reward function, and $\gamma \in (0, 1)$ is a discount factor.
In state $s_t$, the agent performs action $a_t$ according to policy $\pi(a_t|s_t)$, transitions to state $s_{t+1}$ with probability $p(s_{t+1}|s_t,a_t)$, and receives reward $r_{t+1}$.
The process repeats until episode termination.
The objective is to find an optimal policy $\pi^*(a_t|s_t = s)$ for all $s \in S$ to maximise the expected return $\mathbb{E}_{\pi^*}[ \, \sum^{K-1}_{k = 0}\gamma^{k} r_{t+k+1}|s_t = s ]$, where $K$ is the episode length.

The $Q$-function $q^\pi(s,a)$ quantifies the expected return the RL agent can achieve by taking a specific action $a$ in a given state $s$ and following a policy $\pi$ thereafter. Formally,
\begin{equation}
q^\pi(s,a) = \mathbb{E}_{\pi}[ \, \sum^{K-1}_{k = 0}\gamma^{k} r_{t+k+1}|s_t = s, a_t = a ]. \,
 \label{eq:q}
\end{equation}
For an optimal policy $\pi^*$, $q^* = q^{\pi^*}$.

Tabular $Q$-learning is an off-policy algorithm that estimates $q^*(s, a)$ without knowledge of the transition function $p$~\citep{watkins1992q}.
The $Q$-values are learnt through environment interaction and are guaranteed to converge to $q^*(s, a)$ under infinite exploration of all state--action pairs.
The algorithm initialises a $Q$-table randomly for all $(s, a)$. The $Q(s, a)$ values are then updated at each iteration $i$:
\begin{equation}
 Q_{i+1}(s, a) \xleftarrow{\alpha} r(s, a, s') + \gamma \max_{a'} Q_{i}(s',a'),
 \label{eq:q-learning}
\end{equation}
where $\alpha$ is a learning rate.
The notation $x \xleftarrow{\alpha} y$ is expanded as $x \leftarrow x + \alpha(y - x)$.

DDQN~\citep{van2016deep} extends $Q$-learning to large or continuous state spaces with discrete actions, and TD3~\citep{fujimoto2018td3} extends it to continuous action spaces, both at the cost of global optimality guarantees~\citep{mnih2015human}.

\subsection{Boolean Reward Machines}\label{sect:rms}

A Boolean RM is a finite-state automaton that encapsulates the reward structure of the environment.
Transitions between RM states specify the rewards received by the agent.

\begin{definition}[Boolean reward machine]\label{def:rm-bool}
 A \emph{Boolean} RM is a tuple $\MR^{\text{Bool}}_\psa = \langle U, u_0, F, \delta_u, \delta_r \rangle$ given a set of propositional symbols $\MP$, a set of environment states $S$, and a set of actions $A$.
 In the tuple, $U$ is a finite set of states, $u_0$ is an initial state, $F$ is a finite set of terminal states, $\delta_u$ is a state-transition function such that $\delta_u:U \times 2^\props \to U \cup F$, and $\delta_r$ is a state-reward function such that $\delta_r:U \to [S \times A \times S \to \mathbb{R}]$.
\end{definition}

Following~\citet{icarte2022reward}, a labelling function $L: S \times A \times S \to 2^\mathcal{P}$ maps each environment experience $(s, a, s')$ to the set of true propositions; $\delta_u$ then selects the abstract successor RM state; and $\delta_r$ assigns the reward.
Intuitively, an MDP with RMs (MDPRM) is an MDP defined over the cross-product $\tilde{S} = S \times (U\cup F)$: an MDPRM is a tuple $M_{\MR} = \langle \tilde{S}, \tilde{s}_0, \tilde{A}, \tilde{p}, \tilde{r}, \tilde{\gamma} \rangle$, where
$\tilde{s}_0 \in \tilde{S}$ is an initial state;
$\tilde{A} = A$;
state-transition function $\tilde{p}(\langle s', u'\rangle \mid \langle s, u\rangle, a)$ is $p(s'| s, a)$ if $u' = \delta_u(u, L(s,a,s'))$ and $u \in U$, $p(s'| s, a)$ if $u' = u$ and $u \in F$, and $0$ otherwise;
state-reward function $\tilde{r}(\langle s, u\rangle, a, \langle s', u'\rangle)$ is $\delta_r(u)(s, a, s')$ if $u \notin F$ and $0$ otherwise;
and $\tilde{\gamma} = \gamma$ is a discount factor.
The task formulation with respect to MDPRM is Markovian.
Thus, $Q$-learning (among other RL methods) can be used to solve it.
Optimal-solution guarantees of RL algorithms for MDPRMs are the same as for regular MDPs.

\citet{icarte2022reward} introduced $Q$-learning algorithms for MDPRMs.
One is $Q$-learning with RMs (QRM) over the cross-product $S\times U$.
To exploit the RM reward structure,~\citet{icarte2022reward} proposed QRM with counterfactual reasoning (CRM) and a sub-optimal options-based hierarchical RL algorithm---HRM.
In CRM, synthetic experiences are generated for each RM state per environment interaction.

\section{Three New Types of Reward Machines}

We now formally introduce numeric, agenda, and coupled RMs and compare them to Boolean RMs.

\subsection{Numeric Reward Machines}

Numeric RMs compactly model changes in numeric variables, simplifying the RM design process, as shown in the Delivery example (Figure~\ref{fig:rm-num}).
In this example, the numeric variable counts uncollected boxes---remaining subtasks.
More generally, numeric variables can encode any discrete numeric information, such as distance to a goal or resource usage.
Numeric RMs cannot support continuous numeric variables, though, because the induced Boolean RM would have infinitely many states.
To ensure that the induced Boolean RM is finite, we make the following assumption.

\begin{assumption}
 All numeric variables in numeric RMs are discrete and have finite bounds.
\end{assumption}

We introduce numeric features inspired by qualitative numeric planning~\citep{srivastava2011qualitative}, where a numeric feature signals whether a discrete numeric variable increases or decreases after applying an action.
In our setting, numeric variables give a sense of task progression to the agent, so we restrict them to decrease only.

Each numeric variable $w \in W$ is lower-bounded by a value $w_{\ell}$ that specifies the goal value for that variable.
We map each $w$ to a numeric feature $p_w$ with domain $\mathcal{D}_{p_w} = \{\dec$, $\decbar$, $!\}$.
Value $p_w\dec$ indicates that $w$ has decreased since the previous observation but remains above $w_{\ell}$.
Value $p_w\decbar$ indicates that $w$ has reached $w_{\ell}$.
Value $p_w!$ indicates that $w$ has neither decreased nor is equal to $w_{\ell}$.

\begin{proposition} Domain $\mathcal{D}_{p_w}$ fully captures possible changes between the current and previous values of $w$: $w_t$ and $w_{t-1}$.
\end{proposition}

\begin{proof}
Since $w$ cannot increase and is lower-bounded by $w_{\ell}$, there are four possible relations between $w_t$, $w_{t-1}$, and $w_{\ell}$:
(1) $w_{t-1} > w_t > w_{\ell}$ maps to $p_w\dec$;
(2) $w_{t-1} > w_t = w_{\ell}$ and 
(3) $w_{t-1} = w_t = w_{\ell}$ map to $p_w\decbar$; and 
(4) $w_{t-1} = w_t > w_{\ell}$ maps to $p_w!$.
The mapping is thus complete.
\end{proof}

Now, we have all the ingredients to define numeric RMs.

\begin{definition}[Numeric reward machine]\label{def:rm-num}
A \emph{numeric RM} is a tuple $\mathcal R^{\text{num}}_{\psa}=\langle U, u_0, F, \delta_u, \delta_r \rangle$ given sets of
Boolean and numeric features $\MP =  \MP_{\text{Bool}} \cup  \MP_{\text{num}}$,
environment states $S$,
and actions $A$.
$U$, $u_0$, $F$, and $\delta_r$ are defined as in a Boolean RM,
and $\delta_u: U \times 2^{\MP_{\text{Bool}}} \times 3^{\MP_{\text{num}}} \to U \cup F$ is a state-transition function.
Each numeric feature is associated with a discrete numeric variable $w \in W$ with a finite domain $\mathcal{D}_w$.
\end{definition}

Features encode observable environment events.
A numeric RM accepts both Boolean and numeric features.
Boolean features are propositional symbols.

\begin{assumption}[]
For any Boolean feature $p_{\text{B}}$ and state $s$, the agent has access to $\interp{p_{\text{B}}}$, where $\llbracket  \rrbracket$ is the Iverson bracket.
\end{assumption}\noindent
This common assumption ensures that the agent's traversal through the RM aligns with actual environment conditions.
Likewise, we assume that the agent has access to the values of each numeric feature $p_w \in \mathcal{D}_{p_w}$.
That is, the agent can observe whether $w$ has decreased since the previous observation and whether the goal $w_\ell$ has been reached.

Numeric RMs offer a compact way to specify tasks, reducing the modelling effort.
They are particularly useful for tasks in which subtasks can be completed in any order and the agent must learn an optimal ordering.
However, learning with numeric RMs has two challenges.
First, numeric RMs do not provide any task decomposition or intermediate guidance, which is an essential property of RMs.
Second, a policy cannot be learnt directly from a numeric RM because a single RM state may correspond to multiple stages of task progression.
For instance, in the example numeric RM (Figure~\ref{fig:rm-num}), the agent can occupy only two RM states before completing the task, regardless of the number of boxes to deliver.
Specifically, different trajectory segments---corresponding to different numbers of undelivered boxes---all map to the same numeric-RM state $u_0$.
This would cause the agent to repeatedly overwrite learnt information for $u_0$.

To address these issues, numeric RMs must be translated to Boolean RMs by unfolding their states with respect to the domains of numeric variables $\mathcal{D}_{w}$.

\subsection{From Numeric to Boolean Reward Machines}

To address the agent's inability to learn from a user-provided numeric RM, it is translated to a Boolean RM under the following assumptions.

\begin{assumption}[]\label{ass:dom}
 Numeric variables in a numeric RM have finite domains that correspond to the
 collections of different tasks given to the agent.
\end{assumption} \noindent
To enable the functionality of subtask completion in any order, we specifically assume the following:
\begin{assumption}[]
 Sets of tasks encapsulated in numeric variables can be completed in any order.
\end{assumption}
Encapsulation of \emph{ordered} tasks into a \emph{numeric} variable is a useful extension to be realised in future work.
For now, we assume that each subtask in the collection of ordered subtasks is assigned a Boolean feature.
Next, we restrict our numeric RMs as follows:
\begin{assumption}[]\label{ass:one-task}
 A numeric RM is formulated such that only one task for each feature $p_w$ can be completed per step in the environment and, upon its completion, $p_w \dec$ or $p_w \decbar$ signals to transition to the next RM state.
 \end{assumption}

If the assumptions are met, a translation algorithm takes a numeric RM $\MR_{\psa}^{\text{num}}$ along with the numeric-variable domains $\MD_{w}$ as input and outputs a Boolean RM $\MR^{\text{Bool}}_{\mathcal{P}SA}$.
The translation algorithm unfolds the RM states with respect to $\mathcal{D}_{w}$ following the transitions in $\mathcal{R}_{\psa}^{\text{num}}$.
In the translated $\MR^{\text{Bool}}_{\mathcal{P}SA}$, each state corresponds to a collection of subtasks to perform next for each possible completion order.
As a result, the $\MR^{\text{Bool}}_{\mathcal{P}SA}$ representation is exponential in the number of subtasks and can contain symmetric states corresponding to the same $Q$-values.
However, reducing symmetric states requires associating the RM states with the additional information about uncompleted subtasks and completion orders.

Another limitation of Boolean RMs is that CRM is only partially applicable when completed subtasks are removed from the environment, as counterfactual experiences for their associated RM states become incorrect.
For example, in the toy task depicted in Figure~\ref{fig:delivery}, collecting box~$1$ transitions the agent to $u_1$ in the Boolean RM (Figure~\ref{fig:rm-bool}) and clears its cell, so future visits no longer make Boolean feature \texttt{b}$_1$ true.
Consequently, counterfactual experiences for states $u_0$ and $u_4$ will be erroneous, and counterfactual reasoning can only be applied to RM states reachable after $u_1$.

\subsection{Agenda Reward Machines}

We handle the problems above by associating the RM states with an \emph{agenda}---the remaining subtasks to complete.

\begin{definition}[Agenda reward machine]\label{def:rm-sem}
 An agenda RM is a Boolean RM with numeric variables $w \in W$ with domains $\mathcal{D}_w = \agenda$ enumerating unordered subtasks and a labelling function $\Xi: U \cup F \to \{\langle d, \tasksubset, x \rangle \mid \tasksubset \in 2^\agenda\}$, where
 $\agenda$ is the set of all subtasks to complete, $\tasksubset \subseteq \agenda$ is the set of remaining subtasks, $d$ is the RM state depth, and $x$ is the current objective.
 The depth $d$ of RM state $u$ is the distance from the initial state of the RM to $u$.
\end{definition}\noindent
We construct a power set $2^\agenda$ from all subtasks assuming that the agent needs to find an optimal completion order.
In $\langle d, \tasksubset, x \rangle$, $x$ can be either $\tasksubset$ or a Boolean feature which must become true in the environment to trigger the transition to the next RM state.
If $x=\tasksubset$, then, upon completion of any subtask from $\tasksubset$, the agent transitions to the next RM state.

This labelling scheme aligns with the RL principle of learning from information contained in future states (see the $Q$-learning update in (\ref{eq:q-learning})).
By tracking remaining subtasks rather than completed ones, the RM maintains the correct direction of information flow for value backpropagation.
Furthermore, we track the RM state depth so that symmetric states get identical labels and can be merged.

\begin{proposition}
 The labelling scheme in an agenda RM uniquely identifies RM states with respect to task progression.
\end{proposition}
\begin{proof}[Proof]
Assume, for the sake of contradiction, that two RM states have identical labels $\langle d, \tasksubset, x \rangle$ but correspond to different stages of task progression.
However, task progression is completely determined by the remaining subtasks $\tasksubset$, current objective $x$, and RM-state depth $d$.
If it were different for the same $\tasksubset$ and $x$, then the $d$ values would be different.
For example, in task \texttt{a}$\to$\texttt{b}$\to$\texttt{a}, $d = 0$ for the first \texttt{a}, and $d = 2$ for the second \texttt{a}.
Therefore, RM states cannot have equal labels at different task-progression stages.
Conversely, if two states have identical task progression, then by definition they share the same $\tasksubset$, $d$, and $x$.
Thus, their labels coincide.
\end{proof}

We translate the given numeric RM into an agenda RM using a similar algorithm to that of numeric-to-Boolean-RM translation.
The translation differs only in the labelling scheme.
While the agenda RM already provides computational advantages via state reduction, converting it to a coupled RM yields even more benefits.

\subsection{Coupled Reward Machines}

In a coupled RM, agenda-RM states are split by subtasks in the current objective $\tasksubset = \set{\task_1, \task_2, \dots, \task_N}$.

\begin{definition}[Coupled reward machine]\label{def:rm-coupled}
For task set $\tasksubset = \set{\task_1, \task_2, \dots, \task_N}$, a \emph{coupled RM} is an agenda RM where states labelled with $\langle d, \tasksubset, \tasksubset \rangle$ are split into $N$ states $\langle d, \tasksubset, \task_1 \rangle$, $\langle d, \tasksubset, \task_2 \rangle$, \dots, $\langle d, \tasksubset, \task_N \rangle$.
The split states remain coupled such that the agent is in all of them concurrently.
\end{definition}

Algorithmically, the agent occupies all coupled-RM states simultaneously while being in a single environment state.
Moreover, as each coupled-RM state corresponds to exactly one subtask, the agent can learn low-level policies for completing individual subtasks.
The agent has access to the RM label and can therefore identify which subtask to perform.

\section{$\boldsymbol{Q}$-learning with Coupled Reward Machines}

We introduce $Q$-learning with coupled RMs (QCoRM), a new task-decomposition algorithm in which the agent learns low-level policies for completing individual subtasks and a high-level policy for determining their completion order.

\subsection{High-Level Policy}

The high-level policy selects the next subtask from the remaining set $\tasksubset = \{\task_1,\dots,\task_N\}$ by treating the available choices as arms in a stochastic multi-armed bandit~\citep{sutton2018reinforcement}.
Because costs change as low-level policies improve (non-stationarity), we maintain exponentially recency-weighted estimates via a constant step-size incremental update.
Let $\eta(u)$ be the estimated number of steps from each RM state $u$ to a goal.
After each episode $e$, for every RM state $u_e \in U_e$ visited during that episode, we update
$\eta(u_e) \xleftarrow{\smash[b]{\alpha_{\eta}}} \EpiL_{u_e \to \text{goal}}$,
where $\alpha_{\eta}$ is a constant learning rate and $\EpiL_{u_e \to \text{goal}}$ is the observed number of steps from $u_e$ to the goal.
We store $\eta(u)$ in a lookup table with $|U|$ entries, which can grow with the agenda structure, but each update is a single scalar operation.
When exploiting, from the set of coupled states, the agent selects the transition with the lowest cost as the next subtask. 
Once the agent converges to an optimal solution, the stored $\eta(u)$ values equal the optimal steps-to-go, and the high-level policy yields a trajectory of length $\EpiLBest$.
To force exploring all transitions in the RM, we introduce an exploration probability $\xi \in (0,1)$.
During exploration, the agent randomly selects RM transitions with a bias towards unused transitions.

\subsection{Low-Level Policies}

Each state in a coupled RM is associated with exactly one subtask, allowing the agent to learn a separate low-level policy for each subtask in $U' = \agenda \cup B$, where $\agenda$ is the complete agenda and $B$ is the set of all Boolean objectives.
Consequently, the state space of the corresponding MDPRM can be constructed as $S \times U'$ using the set of all subtasks $U'$ rather than the full set of RM states $U$.
The amount of low-level information to learn thus increases only linearly with $|U'|$.
In contrast, Boolean RMs require the full state space
$S \times U$, forcing the agent to learn information exponential in the number of unordered subtasks.

As the agent occupies all coupled states $\langle d, \tasksubset, \task_1 \rangle$, $\langle d, \tasksubset, \task_2 \rangle$, $\dots$, $\langle d, \tasksubset, \task_N \rangle$ at once, it can perform $Q$-learning updates in parallel.
As long as the agent remains in the current set of coupled RM states, the $Q$-values for each subtask $\task_k$, $k = 1, \dots, N$ are updated concurrently:
\begin{equation}
\label{eq:lifted-update}
\begin{aligned}
 Q_{i + 1}(\langle s, \task_k \rangle, a) \xleftarrow{\alpha} \; &r(\langle s, \task_k \rangle, a, \langle s', \task_k \rangle) \;+ \\ &\gamma \max_{a' \in A} Q_{i}(\langle s', \task_k \rangle, a'),
\end{aligned}
\end{equation}
where $\alpha$ is the learning rate,
$\gamma$ is the discount factor,
and $r(\langle s, \task_k \rangle, a, \langle s', \task_k \rangle)$ is the immediate reward.

When the agent completes subtask $\task_{\ell}$ and transitions to the next RM state, the $Q$-values associated with the remaining subtasks $\tasksubset \setminus \task_{\ell}$ are updated according to (\ref{eq:lifted-update}).
The update for RM state $\langle d, \tasksubset, \task_{\ell} \rangle$ is, however,
\begin{equation}
\label{eq:grounding-update}
Q_{i + 1}(\langle s, \task_{\ell} \rangle, a) \xleftarrow{\alpha} R(\EpiL, \task),
\end{equation}
where $\EpiL^{\task_{\ell}} > 0$ is the $\task_{\ell}$ execution duration in timesteps and $R(\EpiL, \task)$ is the final reward for $\task_{\ell}$ completion.

\begin{theorem}[] The final reward that preserves global optimality guarantees of tabular $Q$-learning for any subtask $\task$ is
\begin{equation}
\label{eq:r_K}
R(\EpiL, \task) =
\begin{cases}
\check{R} & \text{if } \delta \EpiL^{\task} = 0, \\
\gamma^{\delta \EpiL^{\task} + 1}\check{R} + \frac{\gamma^{1 - \EpiL^{\task}} - \gamma^{\delta \EpiL^{\task}}}{1-\gamma}r_{\text{mi}} & \text{otherwise},
\end{cases}
\end{equation}
where $\EpiL^{\task}$ is the $\task$ execution duration in timesteps, 
$\check{R} \in \mathbb{R}_{>0}$ is the final reward, $\delta \EpiL^{\task} = \max \set{\EpiLBest^{\task} - \EpiL^{\task}, \EpiL - \EpiLMin}$, 
$\EpiL$ is the episode length,
$\EpiLMin$ is the optimal minimum expected episode length, 
$\EpiLBest^{\task}$ is the optimal expected duration of $\task$ under a globally optimal policy, 
and $r_{\text{mi}} \leq 0$ is the minimum immediate reward.
\end{theorem}

\begin{proof}
\textit{Part 1: Agent prefers globally optimal policies under given $\EpiLBest$ and $\EpiLBest^{\task}$.} Consider an arbitrary subtask $\task$ that is part of some complete policy for the entire task.
The discounted return for completing $\task$ with any \emph{locally sub-optimal} policy should exceed that of any \emph{locally optimal} policy if the former contributes to a \emph{globally optimal} solution.
Let $\pi^{\task}_1$ be a locally optimal but globally sub-optimal policy with the expected local duration $\mathbb{E}[\EpiL^{\task}_1] = \EpiLExp^{\task}_1$ and induced episode length $\EpiLExp_1 > \EpiLBest$.
Let $\pi^{\task}_{\text{op}}$ be a locally sub-optimal but globally optimal policy with the expected local duration $\EpiLBest^{\task}$ and induced episode length $\EpiLBest$.
By assumption, $\EpiLBest^{\task} > \EpiLExp^{\task}_1$.

If the final reward was equal to constant $\check{R}$, the agent would prefer $\pi^{\task}_1$ over $\pi^{\task}_{\text{op}}$, as $\pi^{\task}_1$ is locally optimal and hence yields higher expected return (\ref{eq:q}): $\check{Q}^{\task}_1 > \check{Q}^{\task}_{\text{op}}$ ($\mathbb{E}_{\pi^{\task}_1}[\check{R}\gamma^{\EpiL^{\task}_1 - 1} + \sum_{k_1=0}^{\EpiL^{\task}_1 - 2} \gamma^{k_1} r^{\task}_{k_1+1}] > \mathbb{E}_{\pi^{\task}_{\text{op}}}[\check{R}\gamma^{\EpiL^{\task}_{\text{op}} - 1} + \sum_{k_{\text{op}}=0}^{\EpiL^{\task}_{\text{op}} - 2} \gamma^{k_{\text{op}}} r^{\task}_{k_{\text{op}}+1}]$).
For readability, we omit time index $t$.
To prevent convergence to a globally sub-optimal policy, the final reward depends on $\EpiL^{\task}_1$, and the returns become $Q^{\task}_1 = \mathbb{E}_{\pi^{\task}_1}[R(\EpiL^{\task}_1)\gamma^{\EpiL^{\task}_1 - 1} + \sum_{k_1=0}^{\EpiL^{\task}_1 - 2} \gamma^{k_1} r^{\task}_{k_1+1}]$ and $Q^{\task}_{\text{op}} = \check{Q}^{\task}_{\text{op}}$.
To achieve $Q^{\task}_{\text{op}} > Q^{\task}_1$, the reward must therefore satisfy $R(\EpiLExp^{\task}_1) < \check{R} - \gamma^{1-\EpiLExp^{\task}_1}(\check{Q}^{\task}_1 - \check{Q}^{\task}_{\text{op}})$ for global optimality.

Because in many non-cumulative tasks, agents typically receive non-positive intermediate rewards to avoid non-optimal behaviours~\citep{cui2023reinforcement}, we assume that immediate rewards are bounded between $r_{\text{mi}} \leq 0$ and $r_{\text{ma}} \leq 0$.
We can then bound the return difference as follows:
$\check{Q}^{\task}_1 - \check{Q}^{\task}_{\text{op}} \leq \check{R}(\gamma^{\EpiLExp^{\task}_1 - 1} - \gamma^{\EpiLBest^{\task} - 1}) - \frac{1-\gamma^{\EpiLBest^{\task} - 1}}{1-\gamma}r_{\text{mi}}$.
Rearranging yields the reward bound 
$R(\EpiLExp^{\task}_1) < \gamma^{\EpiLBest^{\task} - \EpiLExp^{\task}_1}\check{R} + \frac{\gamma^{1 - \EpiLExp^{\task}_1} - \gamma^{\EpiLBest^{\task} - \EpiLExp^{\task}_1}}{1-\gamma}r_{\text{mi}}$.
Since $\gamma^{\EpiLBest^{\task} - \EpiLExp^{\task}_1} > \gamma^{\max \set{\EpiLBest^{\task} - \EpiLExp^{\task}_1, \EpiLExp_1 - \EpiLMin} + 1}$, as $\gamma < 1$, expression (\ref{eq:r_K}) satisfies the obtained bound for any local duration $\EpiL^{\task}< \EpiLBest^{\task}$ and any realised episode length $\EpiL > \EpiLMin$.
The reward (\ref{eq:r_K}) uses $\EpiL$ and $\EpiL^{\task}$ available from each episode. 
The return ordering holds almost surely because $Q$-learning averages these into expected $Q$-values~\citep{watkins1992q}.
The result is valid for stationary $\EpiLMin$ and $\EpiLBest^{\task}$, but they are updated as $\EpiLMin \xleftarrow{\smash[b]{\alpha_{\EpiL}}}\EpiL$ and $\EpiLBest^{\task} \xleftarrow{\alpha_{\EpiL}} \EpiL^{\task}$ during the high-level exploitation.

\textit{Part 2: $\EpiLMin$ and $\EpiLBest^{\task}$ converge to $\EpiLMin^*$ and $\EpiLBest^{*\task}$, respectively.}
To push the optimal episode length to minimum, $\EpiLMin$ is updated when $\EpiL \leq \EpiLMin$ during the high-level exploitation, so the sequence $\{\EpiLMin\}$ is non-increasing and bounded below by $\EpiLMin^* > 0$.
At each fixed $\EpiLMin$ and $\EpiLBest^{\task}$, the reward (\ref{eq:r_K}) is stationary; by~\citet{watkins1992q}, $Q$-learning contracts to optimal $Q^*$-values for each low-level policy.
By Part~1, each converged low-level policy achieves $\delta \EpiL^{\task} = 0$, so the reward is the constant $\check{R}$, independent of episode length.
With the reward fixed at $\check{R}$, the Bellman contraction with $\gamma < 1$ independently minimises each subtask duration $\EpiL^{\task}$ toward $\EpiLBest^{*\task}$, since $\gamma^{\EpiL^{\task}-1}\check{R}$ is strictly decreasing in $\EpiL^{\task}$; the global episode length is thus collectively reduced, decrementing $\EpiLMin$ if a shorter episode exists.
Since no episode shorter than $\EpiLMin^*$ exists, the sequence stabilises at $\EpiLMin^*$, the reward permanently fixes at the optimal level, and the globally optimal $Q^*$-values follow.
The local durations $\EpiLBest^{*\task}$ are then fixed and aligned with the optimal episode.
Therefore, the reward (\ref{eq:r_K}) preserves global optimality for any $\EpiL^{\task}$ and $\task$.
\end{proof}

Experiences with transitions between distinct RM states (\ref{eq:grounding-update}) are stored in a dedicated buffer during the episode. 
They are replayed once the episode terminates and the episode length along with the low-level-policy durations is known and the agent exploits, i.e. it selects the best transition based on the lowest $\eta({\langle d, \tasksubset, \task_k \rangle}), k = 1, 2, \dots, N$.
Therefore, the low-level policies are learnt to align with the best high-level policy.
Whenever $\EpiL \leq \EpiLMin$, $\EpiLMin$ is updated by incremental averaging, so that the estimate tracks the minimum episode length.
This global episode-length penalty is needed so that each local policy is aware of the global context and is not greedily optimised for the local duration.

\begin{figure*}[t!]
\centering
\begin{subfigure}[]{0.19\textwidth}
\includegraphics[width=1\textwidth]{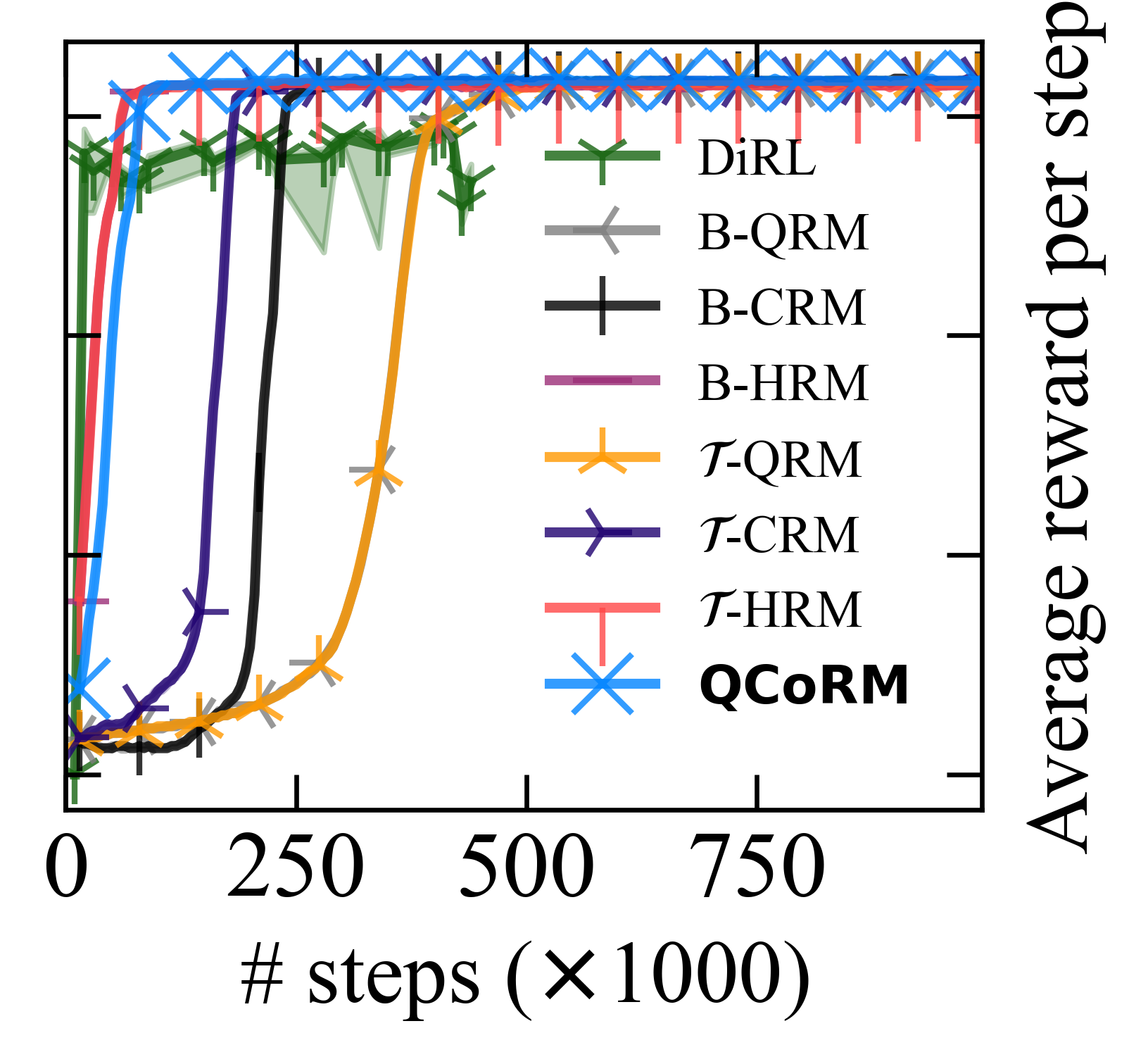}
\caption{Delivery: 2 boxes}\label{fig:delivery-3}
\end{subfigure}
\hfill
\begin{subfigure}[]{0.19\textwidth}
\includegraphics[width=1\textwidth]{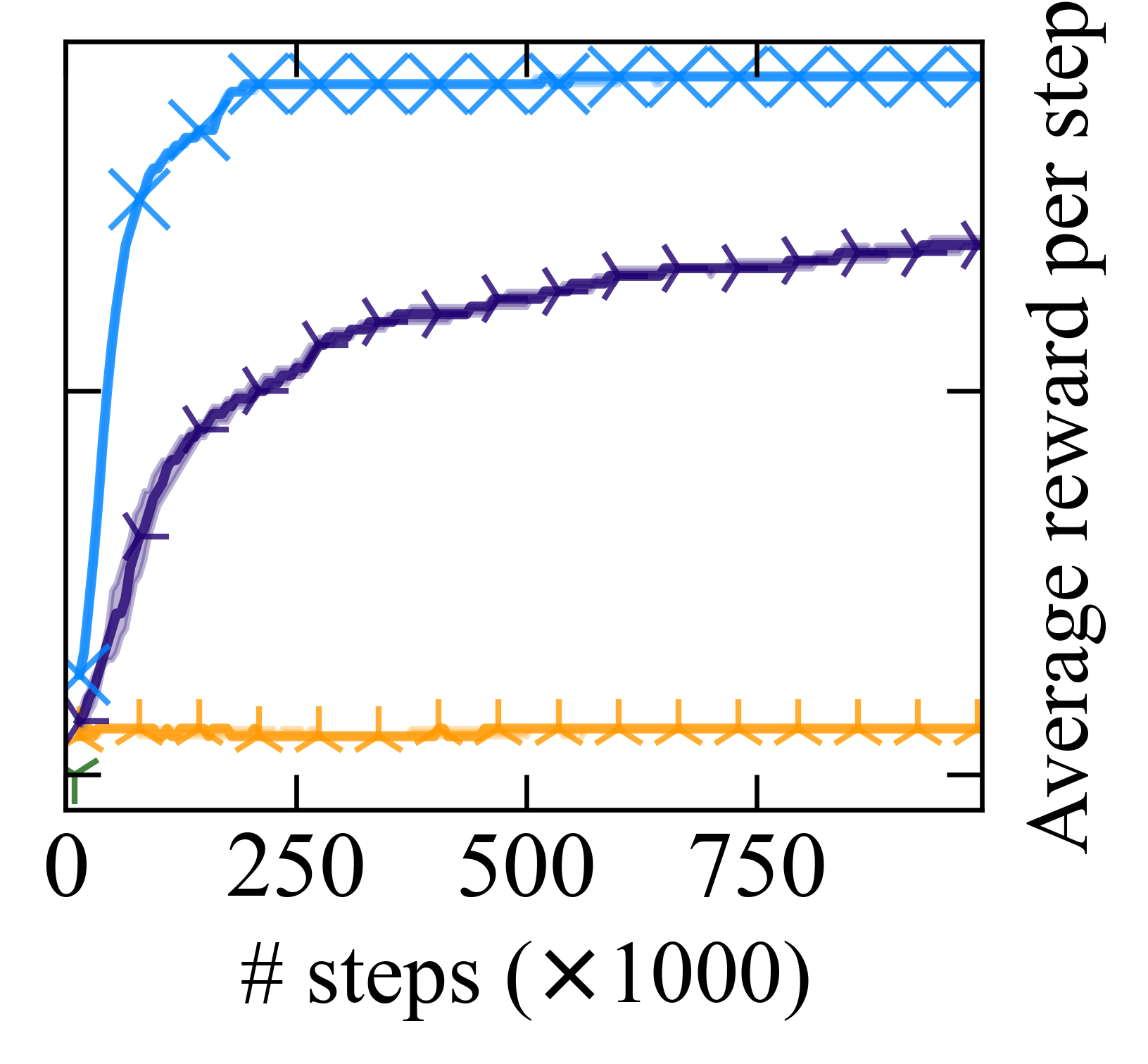}
\caption{Delivery: 8 boxes}\label{fig:delivery-8}
\end{subfigure}
\hfill
\begin{subfigure}[]{0.19\textwidth}
\includegraphics[width=1\textwidth]{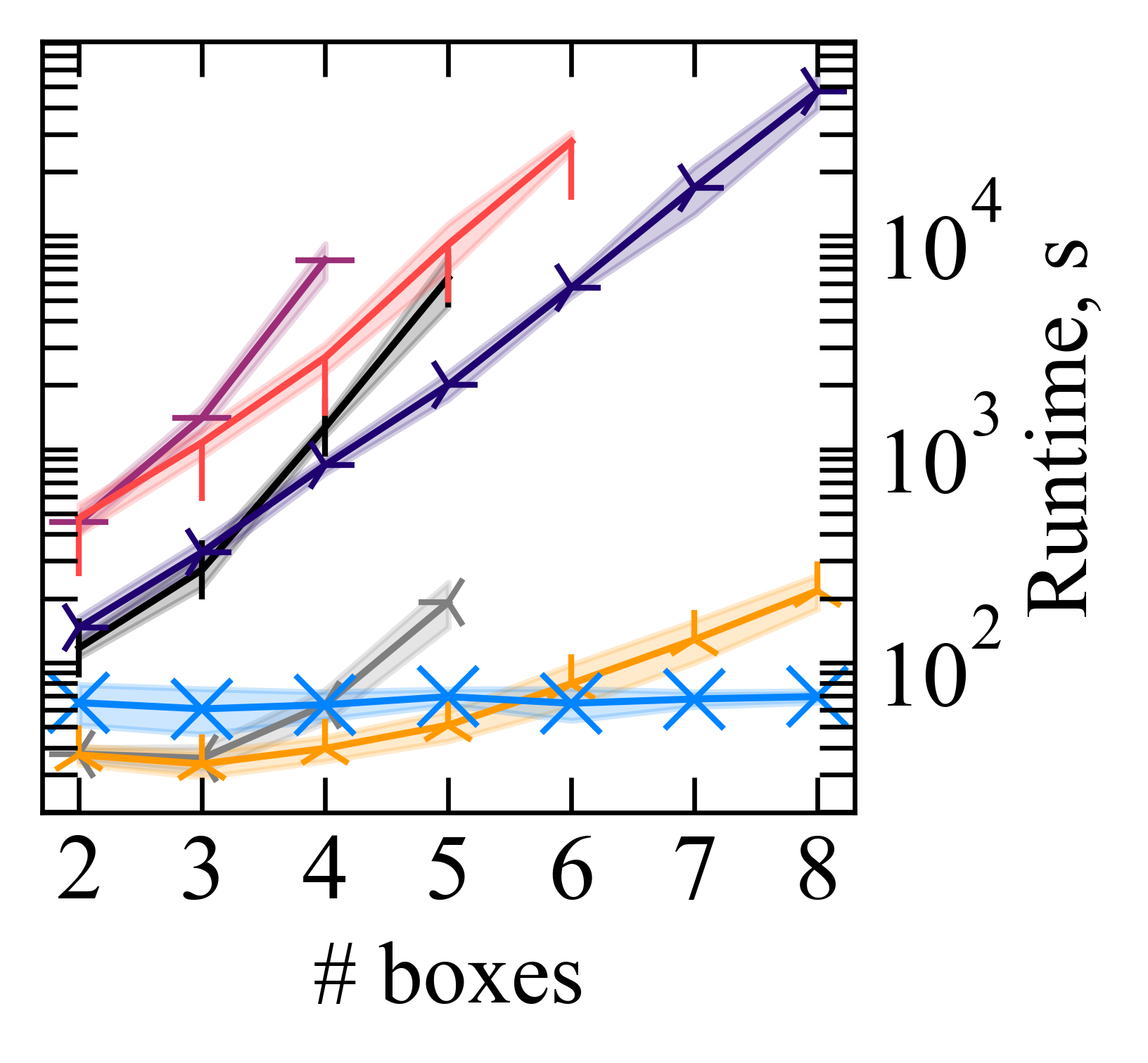}
\caption{Delivery: Runtime}\label{fig:delivery-runtime}
\end{subfigure}
\hfill
\begin{subfigure}[]{0.19\textwidth}
\vspace{2pt}
\setlength{\abovecaptionskip}{21pt}
\begin{tikzpicture}[scale=0.27]
    \draw[gray] (0,0) grid (12,9); 
    \draw[line width=0.5mm] (0,0) rectangle (12,9); 
    \draw[line width=0.5mm] (0,6) -- (1,6); 
    \draw[line width=0.5mm] (2,6) -- (4,6); 
    \draw[line width=0.5mm] (5,6) -- (7,6); 
    \draw[line width=0.5mm] (8,6) -- (10,6); 
    \draw[line width=0.5mm] (11,6) -- (12,6); 

    \draw[line width=0.5mm] (0,3) -- (1,3); 
    \draw[line width=0.5mm] (2,3) -- (4,3); 
    \draw[line width=0.5mm] (5,3) -- (7,3); 
    \draw[line width=0.5mm] (8,3) -- (10,3); 
    \draw[line width=0.5mm] (11,3) -- (12,3); 

    \draw[line width=0.5mm] (3,0) -- (3,1); 
    \draw[line width=0.5mm] (3,2) -- (3,4); 
    \draw[line width=0.5mm] (3,5) -- (3,7); 
    \draw[line width=0.5mm] (3,8) -- (3,9); 

    \draw[line width=0.5mm] (6,0) -- (6,1); 
    \draw[line width=0.5mm] (6,2) -- (6,4); 
    \draw[line width=0.5mm] (6,5) -- (6,7); 
    \draw[line width=0.5mm] (6,8) -- (6,9); 

    \draw[line width=0.5mm] (9,0) -- (9,1); 
    \draw[line width=0.5mm] (9,2) -- (9,4); 
    \draw[line width=0.5mm] (9,5) -- (9,7); 
    \draw[line width=0.5mm] (9,8) -- (9,9); 

    {\scriptsize
    \node at (4.6,4.5) {\texttt{o}$_1$};
    \node at (7.6,4.5) {\texttt{o}$_2$};
    \node at (1.6,7.5) {\texttt{o}$_3$};
    \node at (10.6,7.5) {\texttt{o}$_4$};
     \node at (10.6,1.5) {\texttt{o}$_5$};
     \node at (1.6,1.5) {\texttt{o}$_6$};
    }
    \node at (3.5,6.5) {\Coffeecup};
    \node at (8.5,2.5) {\Coffeecup};

    \node[] at (2.5,1.5) {\small \texttt{A}};

    \foreach \x/\y in {1.5/4.3, 10.5/4.3, 4.5/7.3, 4.5/1.3, 7.5/7.3, 7.5/1.3} {
        \node at (\x,\y) {\Large *};
    }
\end{tikzpicture}
\caption{Office map}\label{fig:office}
\end{subfigure}
\hfill
\begin{subfigure}[]{0.19\textwidth}
    \includegraphics[width=1\textwidth]{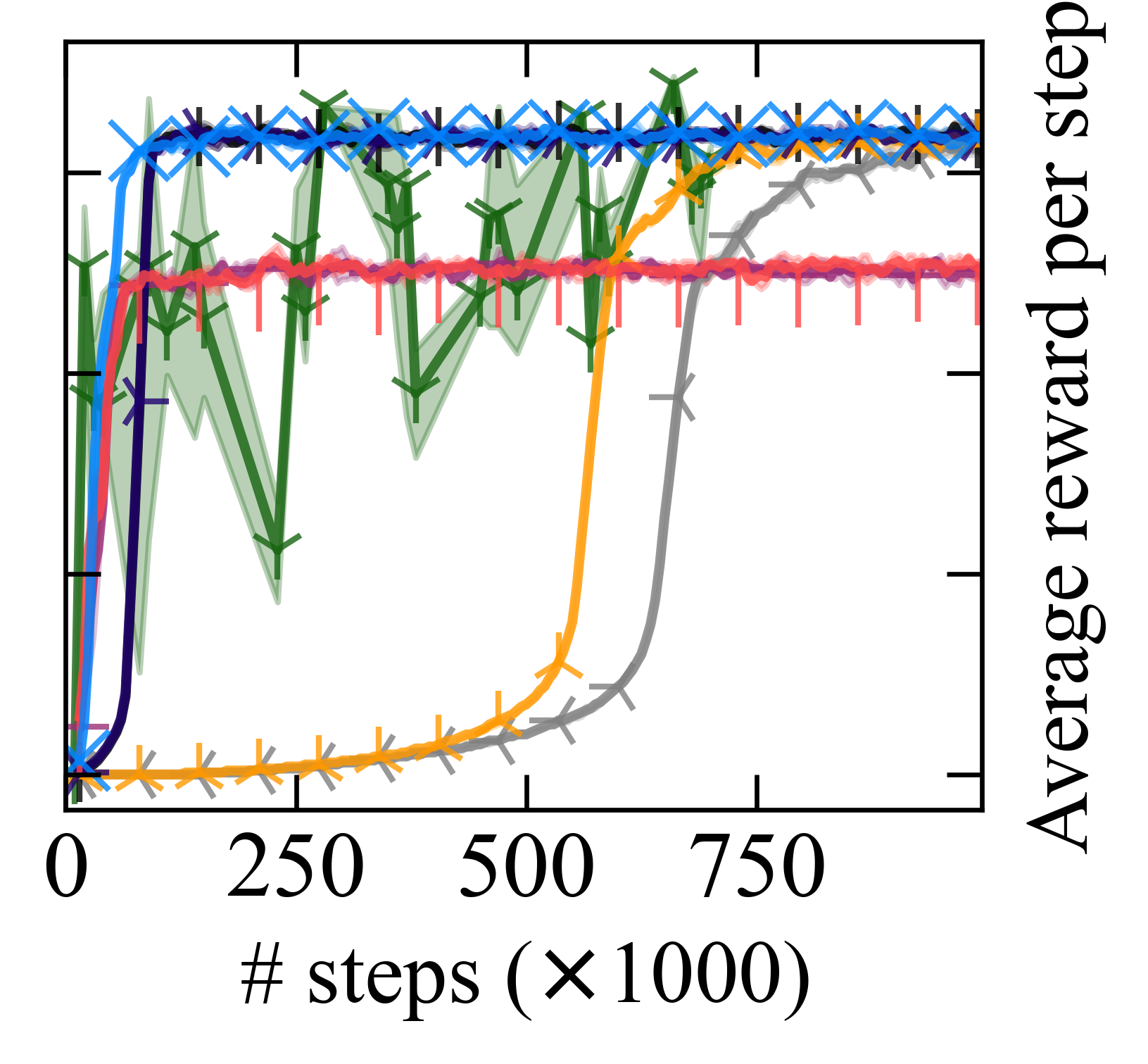}
\caption{Office: 3 offices}\label{fig:office-3}
\end{subfigure}
\vfill
\begin{subfigure}[]{0.19\textwidth}
    \includegraphics[width=1\textwidth]{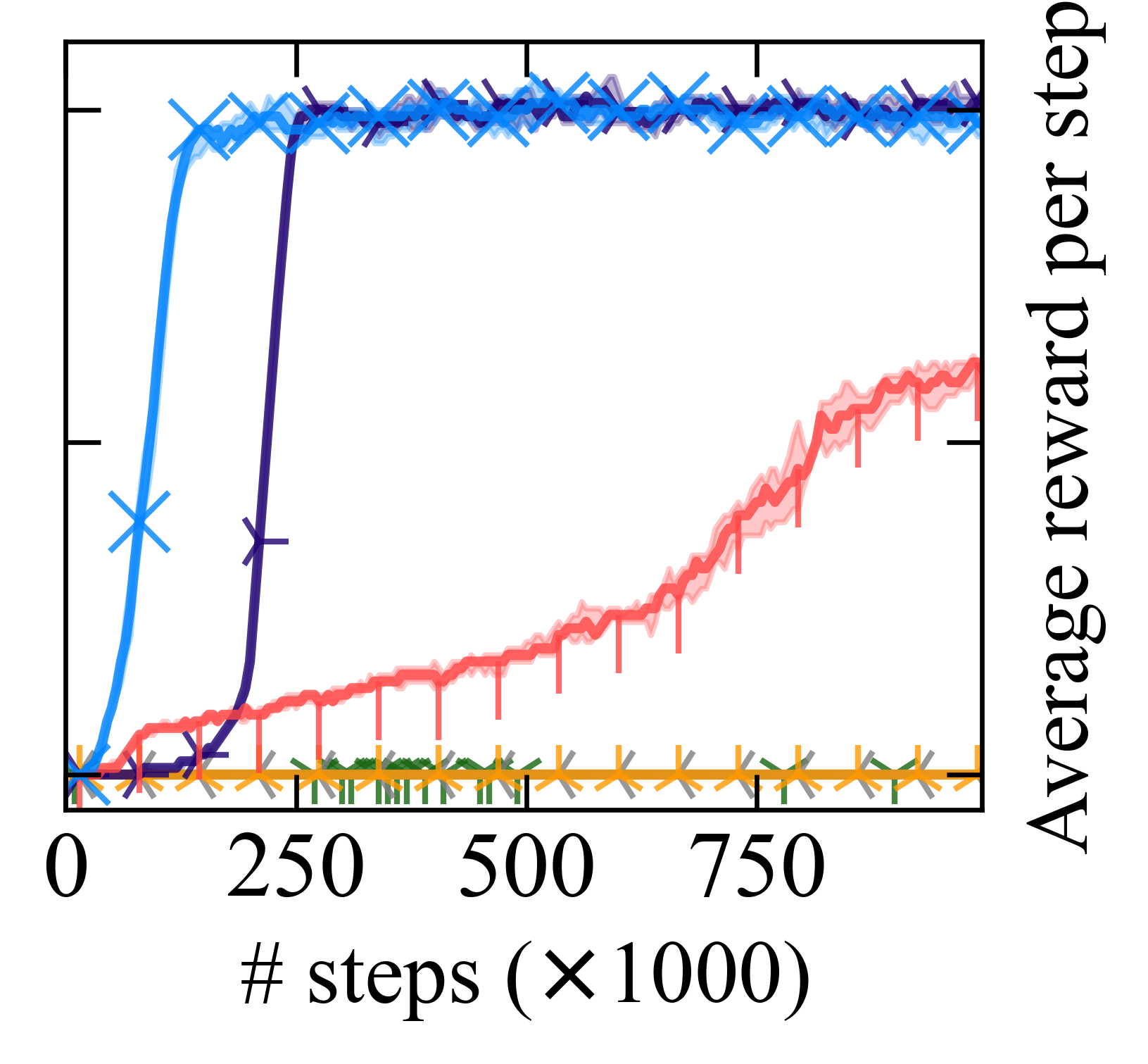}
\caption{Office: 6 offices}\label{fig:office-6}
\end{subfigure}
\hfill
\begin{subfigure}[]{0.19\textwidth}
    \includegraphics[width=1\textwidth]{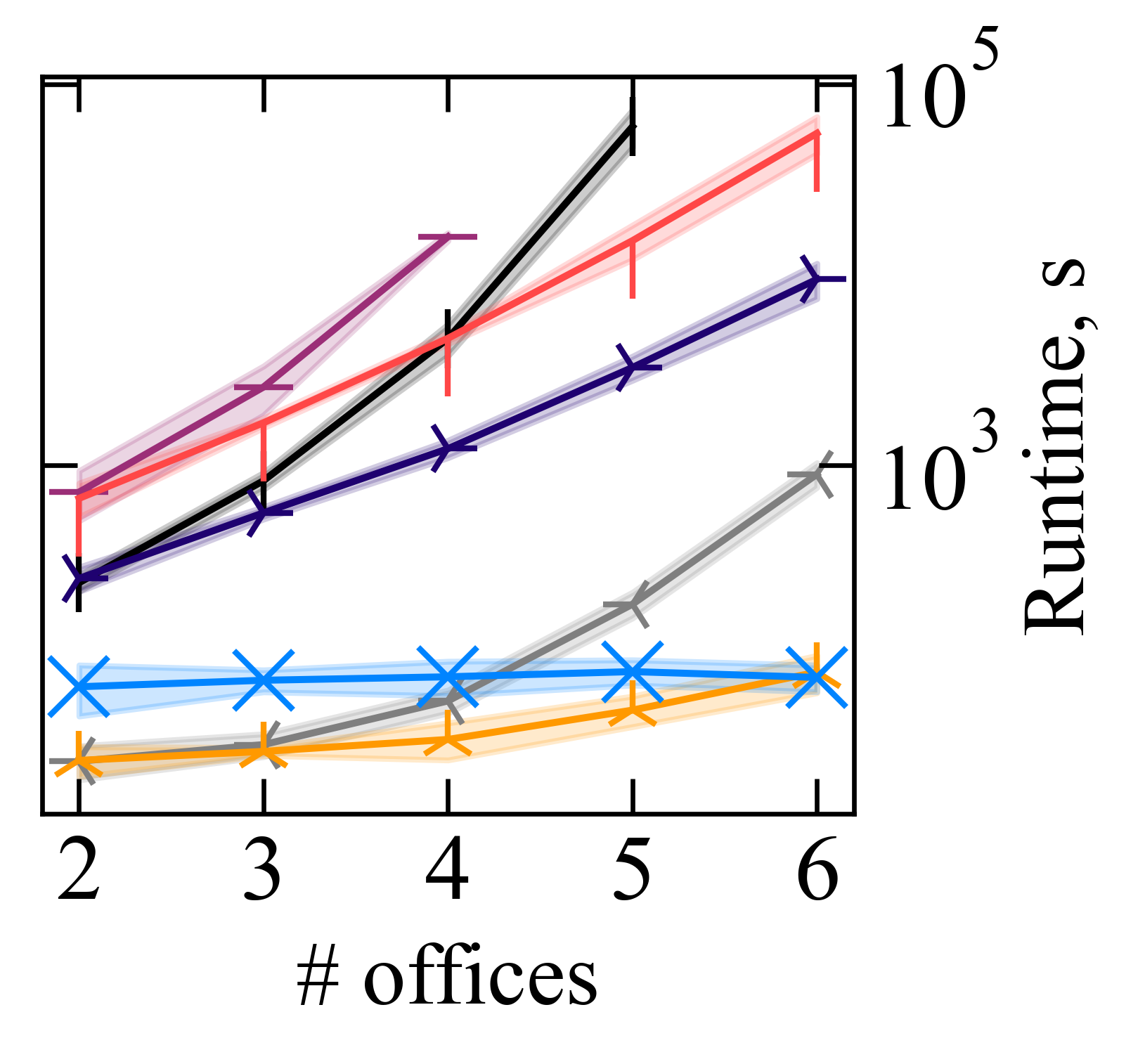}
\caption{Office: Runtime}\label{fig:office-runtime}
\end{subfigure}
\hfill
\begin{subfigure}[]{0.19\textwidth}
\vspace{6pt}
\setlength{\abovecaptionskip}{14pt}
\begin{tikzpicture}[scale=0.4]
    \hspace{0.25cm}
    \draw[thick] (0,0) rectangle (6,6);

    \filldraw[white] (0.5,5.5) circle (0.35);
    \draw[] (0.5,5.5) circle (0.35);

    \foreach \x/\y/\col/\arr in {
        1.8/5.2/pink/45,
        3.5/5.3/yellow!70/15,
        1.2/4.0/cyan/180,
        2.8/4.0/orange!70/165,
        4.0/4.2/yellow!70/67,
        5.2/4.5/cyan/90,
        1.2/2.2/blue!70/45,
        2.3/2.0/blue!70/-45,
        0.8/1.0/olive/-20,
        4.5/2.0/olive/30,
        5.5/1.0/orange!70/-98,
        5.5/2.8/pink/130
    } {
        \filldraw[\col] (\x,\y) circle (0.35);
        \draw[] (\x,\y) circle (0.35);
        \draw[->] (\x,\y) -- ++(\arr:0.7);
    }
\end{tikzpicture}
\vspace{0.2cm}
\caption{Water Domain}\label{fig:water}
\end{subfigure}
\hfill
\begin{subfigure}[]{0.19\textwidth}
    \includegraphics[width=1\textwidth]{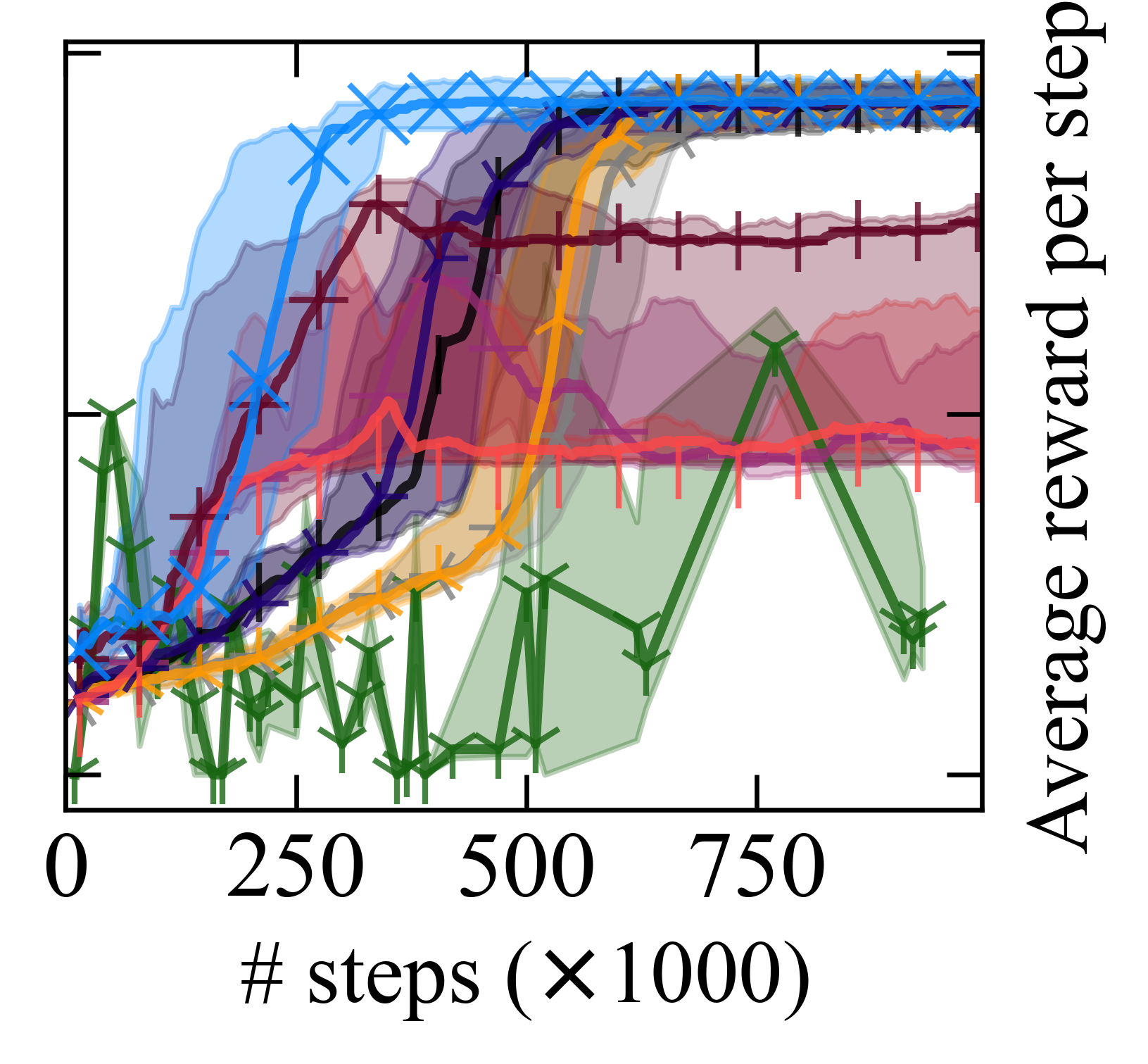}
\caption{Water: 3 balls}\label{fig:water-3}
\end{subfigure}
\hfill
\begin{subfigure}[]{0.19\textwidth}
    \includegraphics[width=1\textwidth]{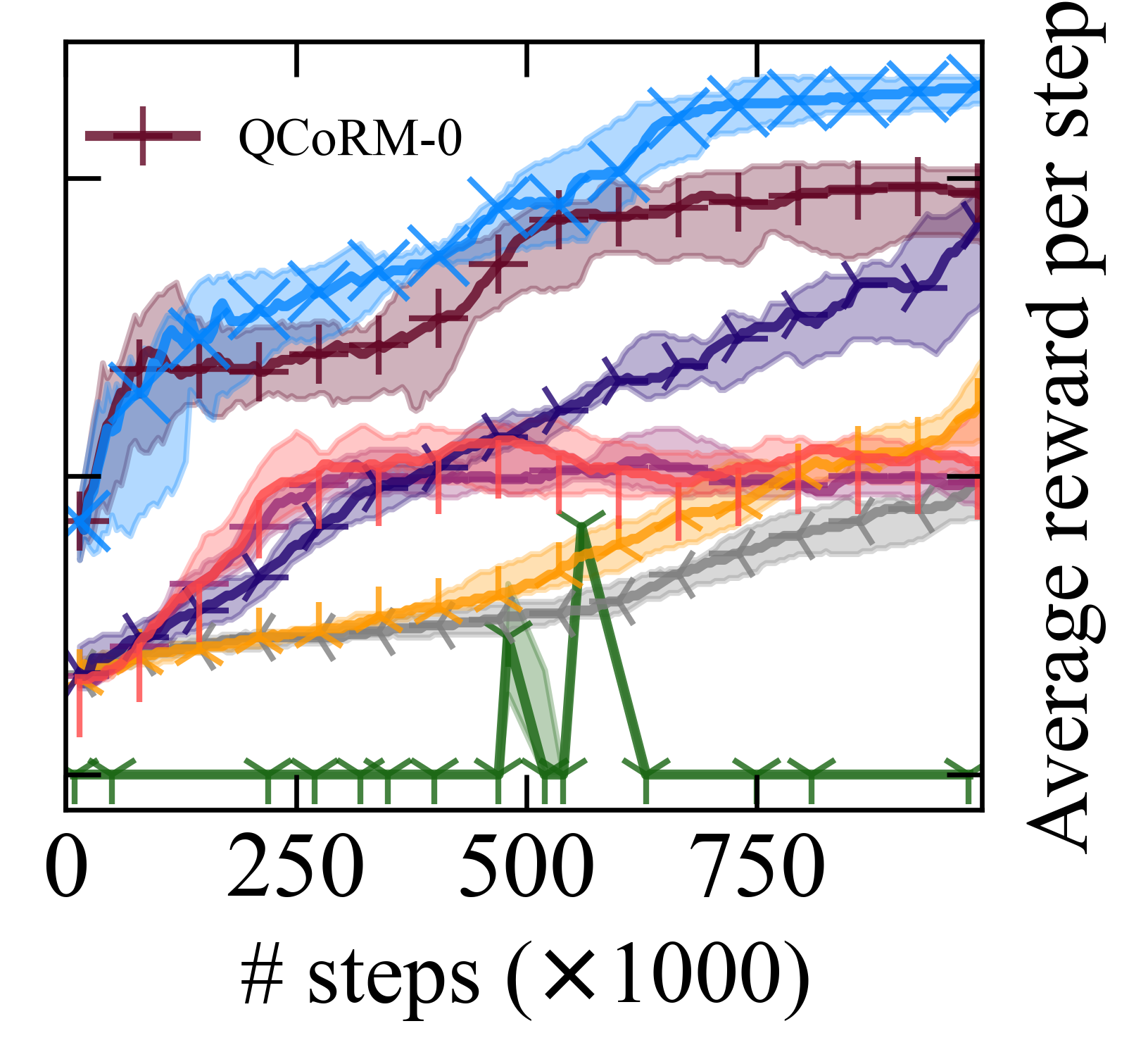}
\caption{Water: 5 balls}\label{fig:water-5}
\end{subfigure}
\vfill
\begin{subfigure}[]{0.19\textwidth}
    \includegraphics[width=1\textwidth]{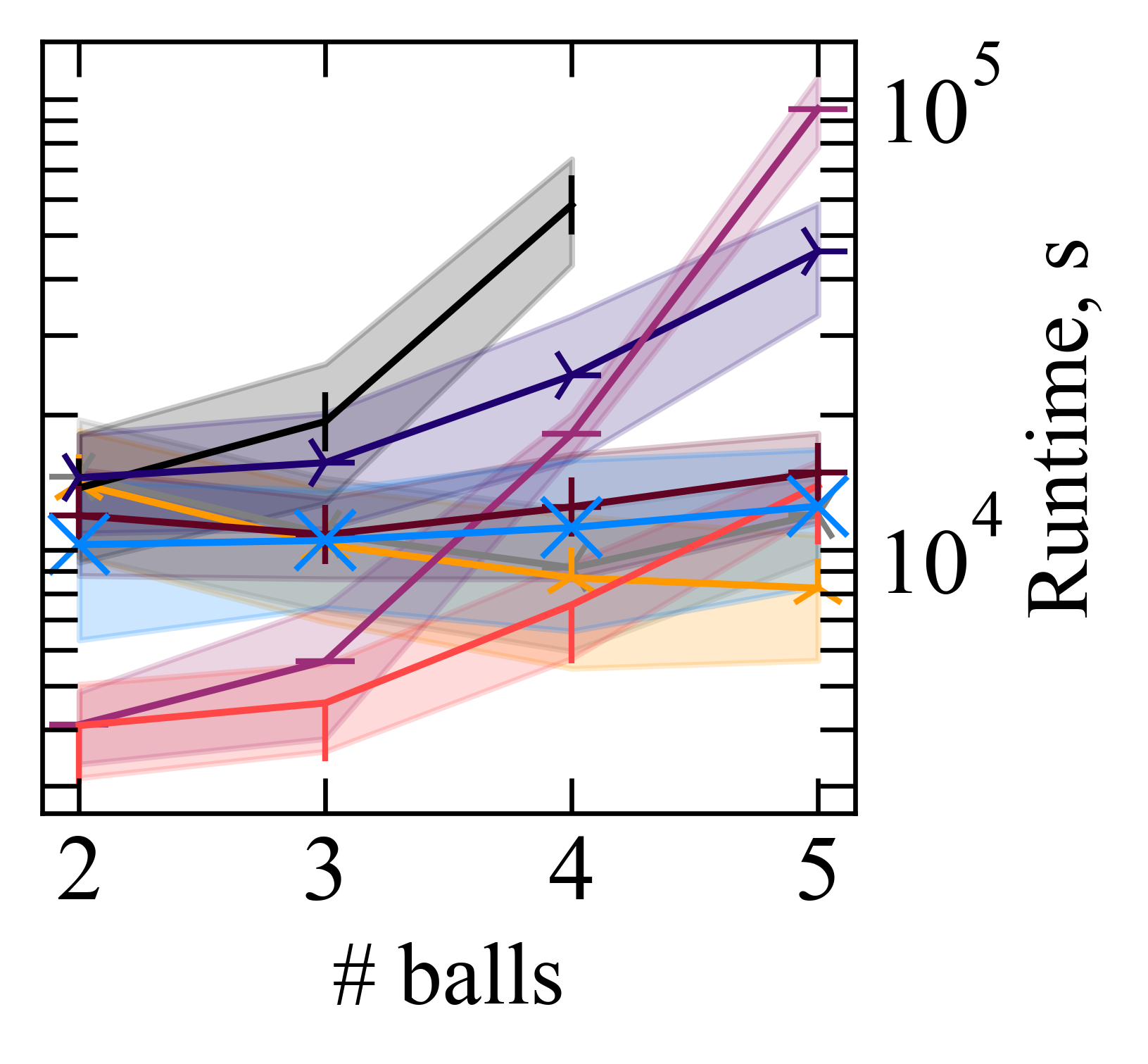}
\caption{Water: Runtime}\label{fig:water-runtime}
\end{subfigure}
\hfill
\begin{subfigure}[]{0.19\textwidth}
\vspace{4pt}
\setlength{\abovecaptionskip}{9pt}
\begin{tikzpicture}
    \node[inner sep=0] (cheetah) at (1.4,1.2) {\includegraphics[width=2.7cm]{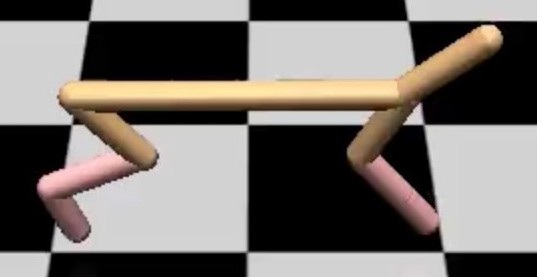}};

    \begin{scope}[yshift=-0.5cm, scale=0.35]
        \draw[-] (-0.5,0) -- (8.5,0);

        \foreach \x/\label in { 0/B, 4/C, 6/D, 8/E} {
            \draw[thick] (\x,-0.2) -- (\x,0.2);
            \node[below] at (\x,-0.3) {\tiny\label};
            \draw[dashed, gray!50] (\x,0.2) -- (\x,1.5);
        }

        \def\agentpos{2}
        \filldraw[orange] (\agentpos,0) circle (5pt);
        \node[above] at (\agentpos,0.1) {\small \texttt{A}};

    \end{scope}
\end{tikzpicture}
\caption{HalfCheetah Domain}\label{fig:half-cheetah}
\end{subfigure}
\hfill
\begin{subfigure}[]{0.19\textwidth}
    \includegraphics[width=1\textwidth]{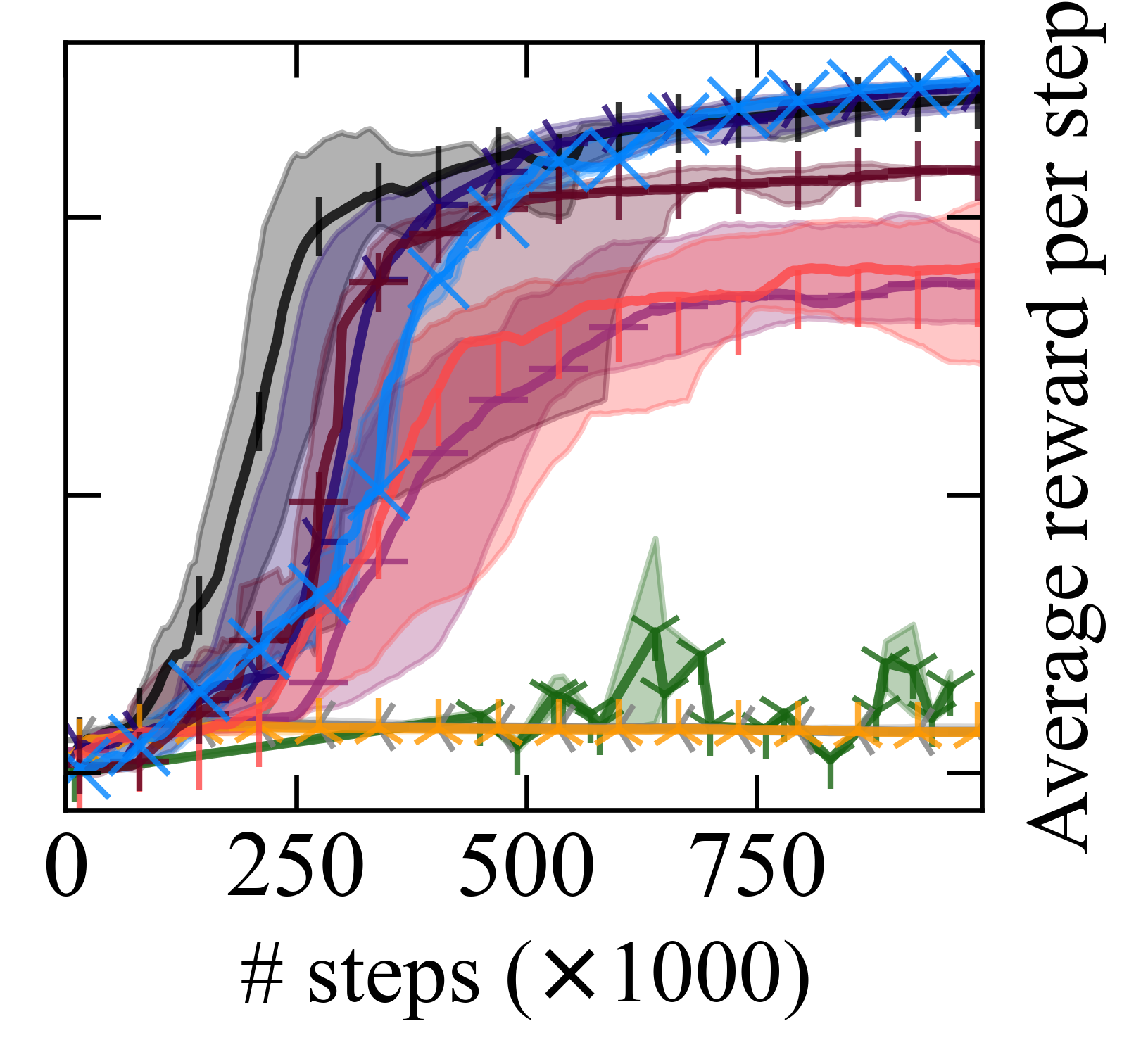}
\caption{HalfCheetah: B,C,D}\label{fig:half-cheetah-3}
\end{subfigure}
\hfill
\begin{subfigure}[]{0.19\textwidth}
\includegraphics[width=1\textwidth]{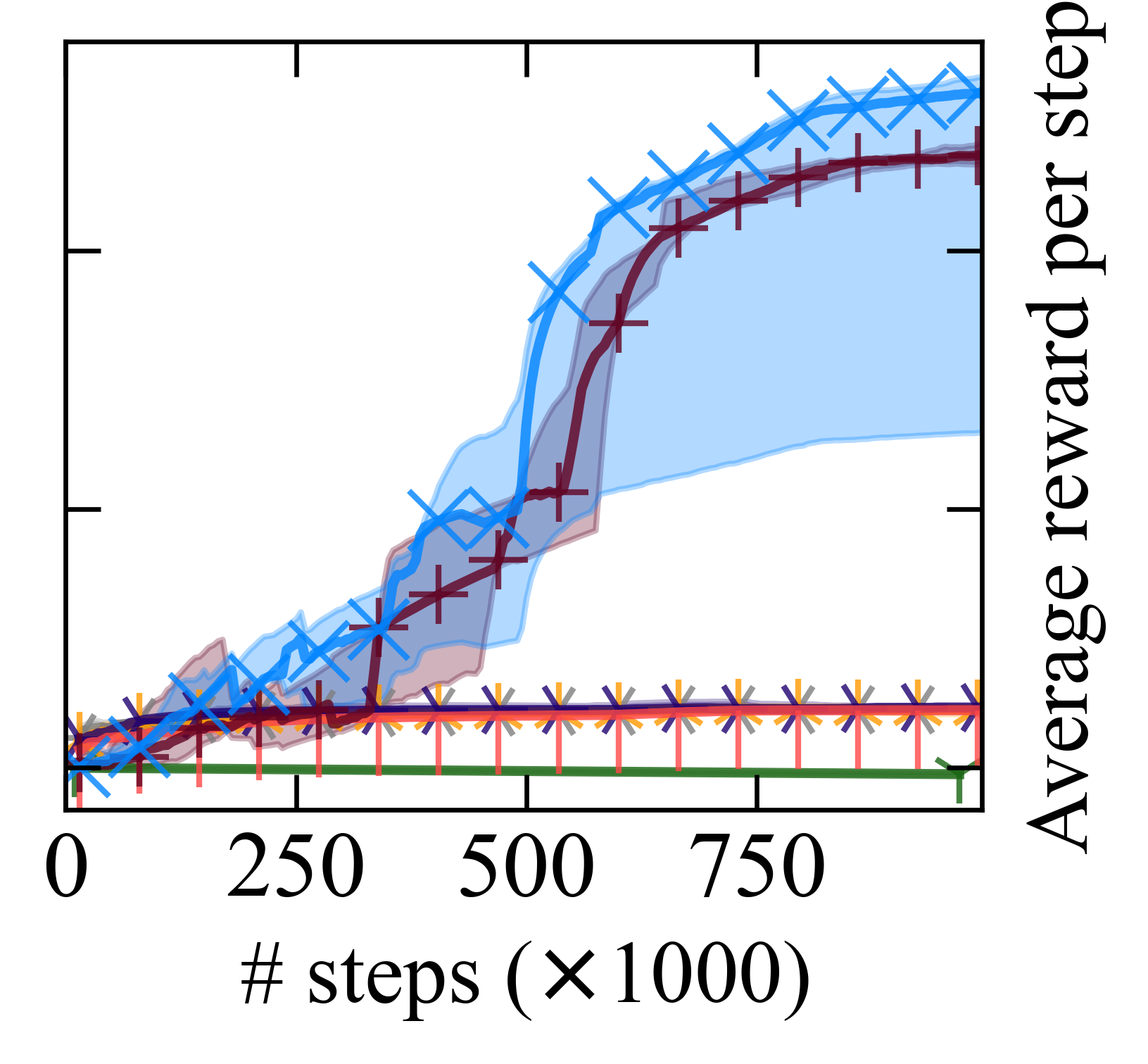}
\caption{HalfCheetah: B,C,D,E}\label{fig:half-cheetah-4}
\end{subfigure}
\hfill
\begin{subfigure}[]{0.19\textwidth}
    \includegraphics[width=1\textwidth]{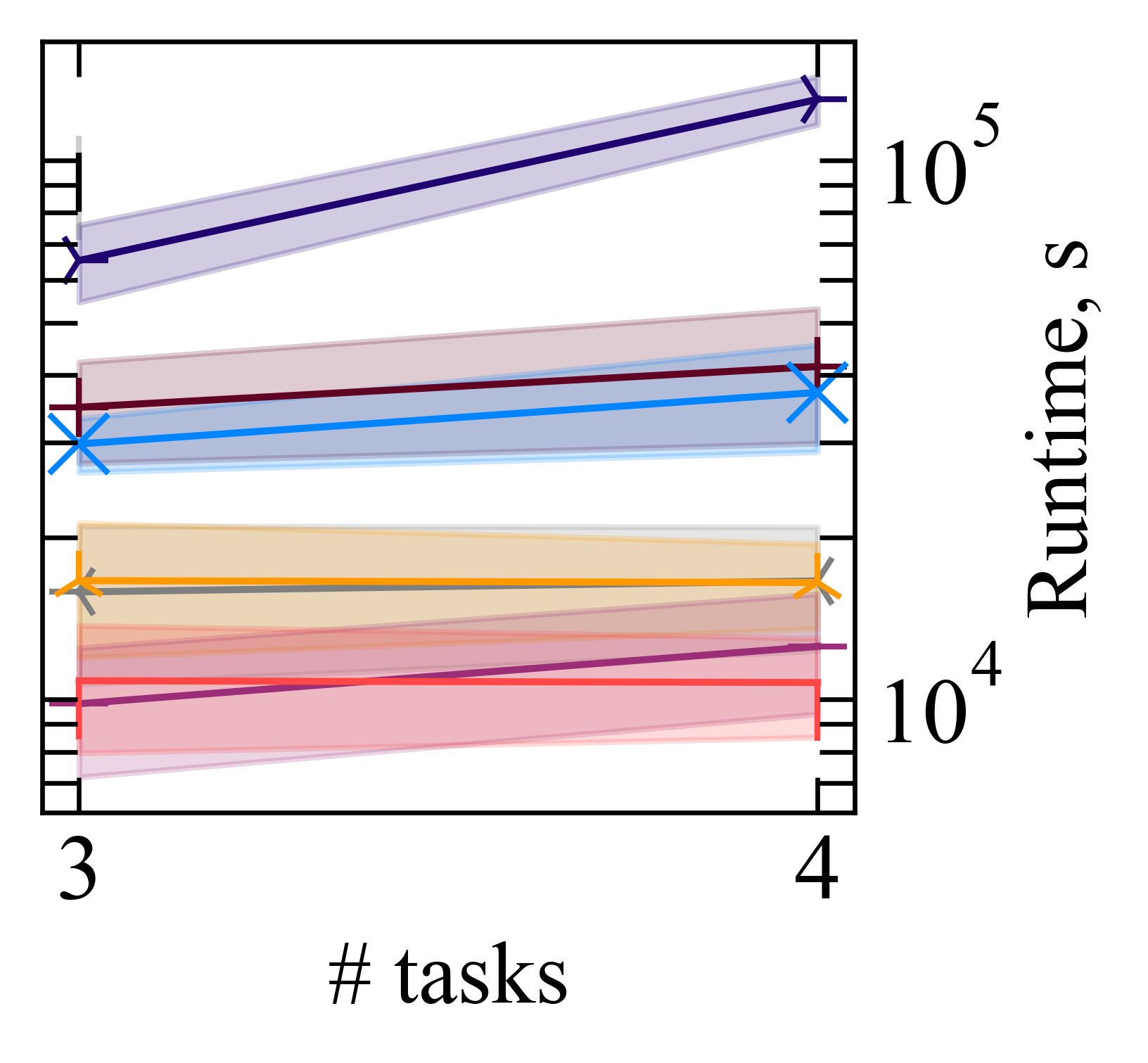}
\caption{HalfCheetah: Runtime}\label{fig:half-cheetah-runtime}
\end{subfigure}
\caption{Results of QCoRM (ours) and baselines, $Q$-learning (QRM), QRM with counterfactual reasoning (CRM), hierarchical HRM, and DiRL. Boolean RMs (B-QRM, B-CRM, B-HRM) and our agenda RMs (\agenda-QRM, \agenda-CRM, and \agenda-HRM) were tested for Delivery (a,b), Office (e,f), Water (i,j), and HalfCheetah (m,n).
As DiRL does not allow setting the number of environment interactions in advance and provides a complete evaluation only after training finishes, we cannot plot DiRL's learning curve.
Instead, we run DiRL with an increasing number of iterations and mark its final performance as a single point per run against the total environment interactions consumed---summing both low-level policy training and high-level policy construction.
QCoRM-0 stands for QCoRM without the joint optimisation from (\ref{eq:r_K}).
(c,g,k,o) Time (in seconds) to run $10^6$ steps in relation to the number of unordered objectives.
(d) Office environment with agent \texttt{A}, offices \texttt{o}$_i, i=1, 2, \dots, 6$, decorations marked by stars, and coffee machines.
(h) Water environment with the agent (white ball) and balls of different colours with arrows indicating their velocities.
(l) HalfCheetah environment with agent \texttt{A} (orange dot) and objectives B, C, D, and E along the horizontal axis.
}
\end{figure*}

\section{Experiments}

We integrate QCoRM with DDQN and TD3 by adding a dedicated replay buffer that stores valuable experiences corresponding to the transitions between different RM states.
After each episode, these stored transitions are replayed to train the network using the same update procedure as above.

Our implementation builds on the code by~\citet{icarte2022reward} and Stable-Baselines3~\citep{raffin2021stable}.
For evaluation, we use two tabular domains, Delivery (our running example) and Office~\citep{icarte2022reward}, and two continuous domains, Water~\citep{karpathy2015reinforcejs} and MuJoCo HalfCheetah~\citep{todorov2012mujoco}.
The agent's initial position is fixed throughout the experiments.
Experiments are run on Intel Xeon Gold 6130 CPUs, using a single core with 16 GiB of memory for the tabular domains and 32 GiB for the continuous domains.
Each algorithm run is limited to $10^6$ steps and $40$ hours.
We report median performance across $10$ seeds, with the shaded regions in plots representing the $25$th and $75$th percentiles.

We compare QCoRM with state-of-the-art RM methods (CRM, HRM, and QRM) and DiRL~\citep{jothimurugan2021compositional}, a compositional learning approach from specifications.
We use the DiRL code from the authors' repository and translate agenda RMs into abstract reachability graphs required by DiRL.
For a fair comparison, we use the same low-level policy learners for DiRL and QCoRM.

We adopt a sparse reward setting, where the agent receives a positive reward only after task completion and zero or negative reward otherwise, because giving positive intermediate rewards can lead to sub-optimal policies~\citep{cui2023reinforcement}.
In particular, for the tabular and Water domains, the agent receives a reward of $1$ after task completion and $0$ otherwise.
For HalfCheetah, the agent receives a reward of $1\,000$ upon completing each subtask and a control penalty otherwise.

\textbf{Parameters.} For tabular $Q$-learning, the learning rate $\alpha = 10^{-5}$, discount factor $\gamma = 0.9$, and exploration parameter $\epsilon = 0.1$.
For DDQN, $\alpha = 10^{-5}$, $\gamma = 0.9$, soft update parameter $\tau = 0.1$, and $\epsilon$ decays linearly from $1$ to $0.1$ in increments of $0.01$.
The replay buffer and batch sizes are $50$K and $32$ for QRM and are scaled up by $|U'|$ for QCoRM and $|U|$ for HRM and CRM.
The RM-exploration parameter $\xi$ decays from $1$ to $0.1$ in increments of $0.001$.
For TD3, $\alpha = 3 \times 10^{-4}$, $\gamma = 0.99$, $\tau = 0.005$, buffer size is $10^6$, and batch size is $256$.
All neural networks have three layers with $512$ hidden units each; training frequency and gradient steps are $1$.
The episode-length learning rate $\alpha_{\EpiL}$ is $0.005$ and $0.05$ for tabular and continuous domains, respectively; and $\alpha_\eta = 0.005$ for all domains.
All other parameters use the default setting.

\subsection{Tabular Domains}

Our \textbf{Delivery} tasks are seven $10\times 10$ grids with $2$--$8$ randomly placed boxes that are removed upon collection.
Figures~\ref{fig:delivery-3} and~\ref{fig:delivery-8} show experimental results for two and eight boxes, respectively.
Although HRM converges fastest for two boxes, it fails to converge for eight boxes.
QCoRM converges to an optimal policy for two and eight boxes, followed by the CRM methods.
DiRL converges to a sub-optimal policy for two boxes and fails to converge for eight.
Notably, QCoRM's performance does not deteriorate after the removal of collected boxes.
In contrast, the CRM methods simulate experiences only for reachable RM states, so fewer experiences are available as the agent traverses the RM.
Convergence is faster for CRM with the agenda RMs than with the Boolean RMs because more states can be identified as reachable due to the more informative labelling scheme.
Boolean RMs for tasks with more than five boxes exceed available memory.

The \textbf{Office} domain~\citep{icarte2022reward}, shown in Figure~\ref{fig:office}, features a grid world with walls, doors, and decorations. 
The agent is given five progressively more complex tasks: deliver coffee to $2$--$6$ offices in any order.
The agent fails the episode upon stepping on a decoration.
Figures~\ref{fig:office-3} and~\ref{fig:office-6} show the results for delivering coffee to three and six offices, respectively.
QRM-based methods converge slowest.
DiRL converges for three coffees but fails for six.
HRM converges to sub-optimal policies.
As in Delivery, QCoRM outperforms CRM-based methods.
In contrast, CRM with both the agenda and Boolean RMs converge at similar rates because offices are not removed from the map after the agent's visit, keeping all the RM states reachable for experience simulation.
We obtained similar results for different numbers of offices.

\subsection{Continuous Domains and Ablation Study}

The \textbf{Water} domain~\citep{karpathy2015reinforcejs} is a two-dimensional arena populated by coloured balls that move at fixed speeds and rebound off walls (Figure~\ref{fig:water}).
The agent, represented by a white ball, adjusts its velocity along four cardinal directions via discrete control.
This domain is challenging because the order in which balls are touched affects the low-level policies for reaching each individual ball.
For example, the agent may need to decelerate before touching one ball to optimally position itself for the next.
As ball positions and velocities are continuous, we use DDQN.
We randomly generate maps with six pairs of differently coloured balls and define progressively more difficult tasks to touch $2$--$5$ balls of specified colours in any order without touching other balls.
Figures~\ref{fig:water-3} and~\ref{fig:water-5} show the results for three and five balls, respectively.
QCoRM again converges fastest, followed by CRM, then HRM (sub-optimal), then QRM, and finally DiRL (sub-optimal).
We observe similar results across all Water instances.

We conduct an \textbf{ablation study} to empirically demonstrate that the reward function (\ref{eq:r_K}) in QCoRM promotes convergence to more optimal policies.
To this end, we remove the joint optimisation from (\ref{eq:r_K}) and give a reward of $1$ upon each subtask completion (agent QCoRM-0).
The results in Figures~\ref{fig:water-3} and~\ref{fig:water-5} show that the agent converges to sub-optimal policies without the joint optimisation, as expected, because low-level policies are not shaped by the global episode-length signal.
Thus, the gap between QCoRM and QCoRM-0 isolates the contribution of reward shaping, while the gap between QCoRM-0 and the CRM baselines reflects the structural advantage of the coupled RM representation.

The \textbf{HalfCheetah} domain~\citep{todorov2012mujoco} involves controlling a two-dimensional bipedal robot to reach specific positions along a horizontal axis (Figure~\ref{fig:half-cheetah}).
We define two tasks where the agent must reach three (B, C, D) or four (B, C, D, E) positions in any order.
We use TD3 for continuous control. 
Because QCoRM delays learning from valuable experiences stored in its dedicated replay buffer, it initially converges more slowly than the CRM-based methods, as we observe in the task to reach three positions (Figure~\ref{fig:half-cheetah-3}).
In the previously discussed domains, this was not an issue because of zero intermediate rewards.
In HalfCheetah, however, the agent receives dense negative feedback, making early learning from all experiences more critical.
Nevertheless, reaching four positions (Figure~\ref{fig:half-cheetah-4}) can only be done by QCoRM, showing its superior scalability.
QRM methods and DiRL fail on both tasks.
Notably, QCoRM without joint optimisation learns to reach the leftmost position B without decelerating, resulting in sub-optimal overall solutions.

\subsection{Runtimes}

We compare the runtimes of the tested algorithms against task size in Figures~\ref{fig:delivery-runtime},~\ref{fig:office-runtime},~\ref{fig:water-runtime}, and~\ref{fig:half-cheetah-runtime}.
In tabular domains, QRM runtimes scale exponentially due to increasing $Q$-table sizes.
In continuous domains, however, QRM runtimes do not change much because the neural network sizes are independent of the task size.
The plots show that QCoRM runtimes scale linearly because the number of low-level simulated experiences grows linearly with the number of subtasks.
In contrast, both HRM and CRM runtimes exhibit exponential growth because the number of simulated experiences is proportional to the number of RM states.
Notably, the exponential blowup in the number of Boolean RM states causes B-CRM and B-HRM to exceed time limits in large tasks.
While the CRM- and HRM-based methods are therefore unsuitable for large tasks, QCoRM handles them efficiently.

\section{Related Work}

Beyond the RMs discussed in our paper, a separate line of work uses automata in RL for task decomposition.
In these approaches, the automaton structure is inferred from experience rather than provided by the user.
First,~\citet{furelos2021induction} introduced ISA---a method for discovering subgoal automata with binary rewards ($0$ or $1$) from traces.
Next,~\citet{furelos2023hierarchies} introduced hierarchies of RMs (HRMs), which are learnt from traces.
Compared to the RMs in our work, HRMs only support labelling edges with real-valued rewards rather than reward functions.
Moreover, unlike in QCoRM, the global policy learnt with HRMs is not guaranteed to be optimal because each automaton is learnt in isolation.
More recently,~\citet{ardon2025form} introduced first-order RMs (FORMs), where first-order logic is used for more compact task specifications.
However, FORMs still support only real-valued rewards.
Furthermore, in contrast to the RMs discussed in our paper, for any number of unordered subtasks, there is only one FORM state expressed by a universally quantified FO atom, thus providing no guidance for unordered-subtask completion.

Below, we compare our QCoRM algorithm with task-decomposition RL algorithms for tackling long-horizon tasks composed of unordered subtasks.
RM-based RL belongs to a class of approaches in which tasks are decomposed into symbolically defined subtasks.
By contrast, approaches that focus on discovering low-level skills or options~\citep{bacon2017option,klissarov2023deep} or do not accept symbolic specifications as input~\citep{nachum2018data,mendez2022modular} are outside the scope of this comparison but could be complementary to RMs.

Learning a policy for each subtask is well-established in the literature~\citep{mohan2024structure}.
Some notable works include options~\citep{sutton2018reinforcement}, policy sketches~\citep{andreas2017modular}, taskable RL~\citep{illanes2020symbolic}, and hierarchical policies~\citep{drexler2023learning}.
These approaches typically optimise Markovian returns and do not, by default, provide semantics for non-Markovian rewards.

Supporting history-dependent objectives generally requires state augmentation or an explicit automaton~\citep{skalse2023limitations}.
In addition to RMs, some notable works include R-AVI~\citep{jothimurugan2021abstract}, DiRL~\citep{jothimurugan2021compositional}, and LSTS~\citep{shukla2024logical}.
Both DiRL and LSTS translate SPECTRL tasks into abstract graphs.
A high-level policy for selecting transitions in the graph is then determined using Dijkstra's algorithm with costs based on the transition success probabilities between different state-space regions.

Likewise, QCoRM uses the RM structure to guide high-level decision making.
However, QCoRM converges to optimal solutions via shaping low-level policies by the global episode-length signal.
In contrast, DiRL's local policies optimise only for individual abstract transitions, so the composed high-level policy is not guaranteed to be globally optimal.
Moreover, DiRL learns one low-level policy at a time, whereas QCoRM learns a low-level policy for each subtask in all coupled RM states in parallel.
This distinction becomes crucial when tasks contain many unordered subtasks.

\section{Conclusions}

We introduce three generalisations of RMs to tackle long-horizon tasks with unordered subtasks.
($1$) Numeric RMs assist the user with compact task representation.
($2$) States in agenda RMs are associated with an agenda---the remaining subtasks to complete.
($3$) In coupled RMs, states are split by subtasks in the agenda, allowing for parallel learning of low-level policies and high-level decision-making on top.
We develop a new task-decomposition algorithm for $Q$-learning with coupled RMs (QCoRM) that shows computational advantages over baseline methods in tabular and continuous domains.
In particular, QCoRM scales well for long-horizon tasks with unordered subtasks, while preserving the optimality guarantees of tabular $Q$-learning.
In future work, QCoRM could be extended beyond $Q$-learning-based methods.

\section{Acknowledgments}

This work was supported by Ericsson Research and the Wallenberg AI, Autonomous Systems, and Software Program (WASP) funded by the Knut and Alice Wallenberg Foundation. The computations were enabled by resources provided by the National Academic Infrastructure for Supercomputing in Sweden (NAISS), partially funded by the Swedish Research Council through grant agreement no.~$2022$-$06725$.

\bibliography{aaai2026}

\end{document}